\documentclass[10pt,twocolumn,letterpaper]{article}

\usepackage{cvpr}
\usepackage{times}
\usepackage{epsfig}
\usepackage{graphicx}
\usepackage{amsmath}
\usepackage{amssymb}
 
\usepackage{subcaption} 
\usepackage{multirow}
\usepackage{algorithmic}
\usepackage{tikz}  
\usetikzlibrary{bayesnet}
 
\usepackage[breaklinks=true,bookmarks=false]{hyperref}

\graphicspath{ {images/} }

\newcommand{\bm}[1]{{\mathbf{#1}}}
\newcommand{\rhom} {\boldsymbol{\rho}}
\newcommand{\pim} {\boldsymbol{\pi}}
\newcommand{\sigmam} {\boldsymbol{\sigma}}
\newcommand{\thetam} {\boldsymbol{\theta}}

\cvprfinalcopy

\begin{document}

%%%%%%%%% TITLE
\title{Unsupervised Learning and Segmentation of Complex Activities from Video}

\author{Fadime Sener, Angela Yao \\
University of Bonn, Germany \\ 
{\tt\small \{sener,yao\}@cs.uni-bonn.de} 
}

\maketitle

%%%%%%%%% ABSTRACT
\begin{abstract} 
This paper presents a new method for unsupervised segmentation of complex activities from video into multiple steps, or sub-activities, without any textual input. We propose an iterative discriminative-generative approach which alternates between discriminatively learning the appearance of sub-activities from the videos' visual features to sub-activity labels and generatively modelling the temporal structure of sub-activities using a Generalized Mallows Model. In addition, we introduce a model for background to account for frames unrelated to the actual activities. Our approach is validated on the challenging Breakfast Actions and Inria Instructional Videos datasets and outperforms both unsupervised and weakly-supervised state of the art.
\end{abstract}

%%%%%%%%% BODY TEXT 
\vspace{-0.9cm}
\section{Introduction}
~\label{sec:intro}
We address the problem of understanding complex activities from video sequences. A complex activity is a procedural task with multiple steps or sub-activities that follow some loose ordering. Complex activities can be found in instructional videos; YouTube hosts hundreds of thousands of such videos on activities as common as \emph{`making coffee'} to the more obscure \emph{`weaving banana fibre cloths'}. Similarly, in assistive robotics, a robot that can understand and parse the steps of a household task such as \emph{`doing laundry'} can anticipate and support upcoming steps or sub-activities.
 
Complex activity understanding has received little attention in the computer vision community compared to the more popular simple action recognition task. In simple action recognition, short, trimmed clips are classified with single labels, \eg of sports, playing musical instruments~\cite{karpathy2014large, soomro2012ucf101}, and so on. Performance on simple action recognition has seen a remarkable boost with the use of deep architectures~\cite{karpathy2014large,simonyan2014two,tran2015learning}. Such methods however are rarely applicable for temporally localizing and/or classifying actions from longer, untrimmed video sequences, usually due to the lack of temporal consideration. Even works which do incorporate some modelling of temporal structure~\cite{fernando2015modeling,sharma2015action,srivastava2015unsupervised,tran2015learning} do little more than capturing frame-to-frame changes, which is why the state of the art still relies on either optical flow~\cite{simonyan2014two} or dense trajectories~\cite{tran2015learning,wang2013dense}. Moving towards understanding complex activities then becomes even more challenging, as it requires not only parsing long video sequences into semantically meaningful sub-activities, but also capturing the temporal relationships that occur between these sub-activities.

We aim to discover and segment the steps of a complex activity from collections of video in an unsupervised way based purely on visual inputs. Within the same activity class, it is likely that videos share common steps and follow a similar temporal ordering. To date, works in a similar vein of unsupervised learning all require inputs from narration; the sub-activities and sequence information are extracted either entirely from~\cite{MalmaudCooking,alayrac2016unsupervised}, or rely heavily~\cite{sener2015unsupervised} on text. Such works assume that the text is well-aligned with the visual information of the video so that visual representations of the sub-activity are learned from within the text's temporal bounds. This is not always the case for instructional videos, as it is far more natural for the human narrator to first speak about what will be done, and then carry out the action. Finally, reliably parsing spoken natural language into scripts\footnote{Here, we refer to the NLP definition of script as \emph{``a predetermined, stereotyped sequence of actions that define a well-known situation''}~\cite{schank1975scripts}.} is an unsolved and open research topic in itself. As such, it is in our interest to rely only on visual inputs.

\begin{figure*}[htb]
	\centering
	\begin{subfigure}[b]{0.60\textwidth}
	 	\includegraphics[width=\columnwidth]{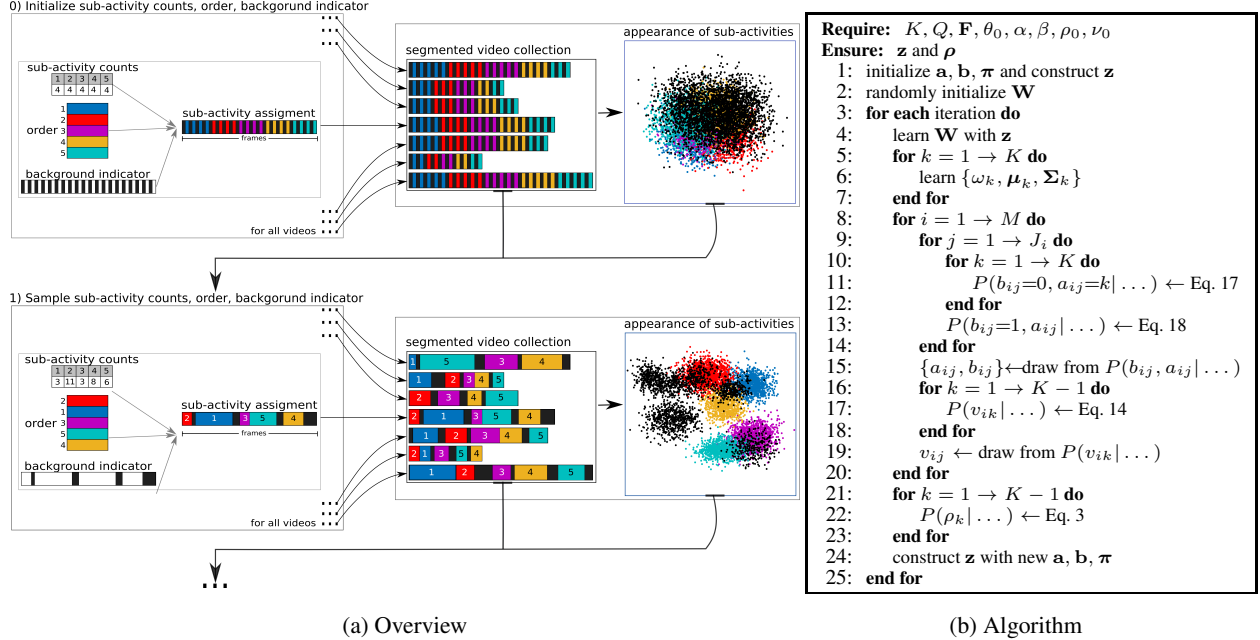} 
		\caption{Overview}
		\label{fig:overviewFigure}
	\end{subfigure}
 \begin{subfigure}[b]{0.32\textwidth}
	 \input{model_alg}
	  \caption{Algorithm}
	  \label{fig:overviewAlgo}
	 \end{subfigure} 
\vspace{-0.3cm}
	\caption{(a) Our iterative model alternates between learning visual appearance and temporal structure of sub-activities. We combine visual appearance with a temporal model to obtain a segmentation of video sequences which is then used to update the visual appearance representation for the next iteration. (b) Algorithm for our model. (Figure best viewed in color.)} 
\label{fig:overview}
\vspace{-0.3cm}
\end{figure*}

In this work, we propose an iterative model which alternates between learning a discriminative representation of a video's visual features to sub-activities and a generative model of the sub-activities' temporal structure. By combining the sub-activity representations with the temporal model, we arrive at a segmentation of the video sequence, which is then used to update the visual representations (see Fig.~\ref{fig:overviewFigure}). We represent sub-activities by learning linear mappings from visual features to a low dimensional embedding space with a ranking loss. The mappings are optimized such that visual features from the same sub-activity are pushed together, while different sub-activities are pulled apart.

Temporally, we treat a complex activity as a sequence of permutable sub-activities and model the distribution over permutations with a Generalized Mallows Model (GMM)~\cite{fligner1986distance}. GMMs have been successfully used in the NLP community to model document structures~\cite{chen2009content} and script knowledge~\cite{frermann2014hierarchical}. In our method, the GMM assumes that a canonical sequence ordering is shared among videos of the same complex activity. There are several advantages of using the GMM for modelling temporal structure. First and foremost, the canonical ordering enforces a global ordering constraint over the activity -- something not possible with Markovian models~\cite{kuehne2014language,richard2017weakly,sener2015unsupervised} and recurrent neural networks (RNNs)~\cite{yeung2016end}. Secondly, considering temporal structure as a permutation offers flexibility and richness in modelling. We can allow for missing steps and deviations, all of which are characteristic of complex activities, but cannot be accounted for with works which enforce a strict ordering~\cite{alayrac2016unsupervised}. Finally, the GMM is compact -- parameters grow linearly with the number of sub-activities, versus quadratic growth in pairwise relationships, \eg in HMMs. 

Within a video, it is unlikely that every frame corresponds to a specified sub-activity; they may be interspersed with unrelated segments of actors talking or highlighting previous or subsequent sub-activities. Depending on how the video is made, such segments can occur arbitrarily. It becomes difficult to maintain a consistent temporal model under these uncertainties, which in turn affects the quality of visual representations. In this paper we extend our segmentation method to explicitly learn about and represent such ``background frames'' so that we can exclude them from the temporal model. To summarize our contributions:
\begin{itemize}
\vspace{-0.23cm}
\item We are the first to explore a fully unsupervised method for temporal understanding of complex activities in video without requiring any text. We design a discriminative appearance learning model to enable the use of GMMs on state-of-the-art visual features~\cite{sanchez2013image,tran2015learning,wang2013dense}.
\vspace{-0.23cm}
\item We verify our method on real-world videos of complex activities which do not follow strict orderings and are heavily interspersed with background frames.
\vspace{-0.23cm}
\item We demonstrate that our method achieves competitive results comparable to or better than the state of the art on two challenging complex activity datasets, Breakfast Actions~\cite{kuehne2014language} and Inria Instructional Videos~\cite{alayrac2016unsupervised}. 
\end{itemize}

\section{Related Work}
~\label{sec:rel_work} 
Modelling temporal structures in activities has been focused predominantly at a frame-wise level~\cite{fernando2015modeling,sharma2015action,srivastava2015unsupervised,tran2015learning}. Existing works on complex activity understanding typically require fully annotated video sequences with start and end points of each sub-activity~\cite{kuehne2014language,richard2016temporal,rohrbach2012database}. Annotating every frame in videos is expensive and makes it difficult to work at a large scale. Instead of annotations, a second line of work tries to use cues from accompanying narrations ~\cite{alayrac2016unsupervised,MalmaudCooking,sener2015unsupervised}. These works assume that the narrative text is well-aligned with the visual data, with performance governed largely by the quality of the alignment. For example, in the work of Alayrac \etal~\cite{alayrac2016unsupervised}, instruction narrations are used as temporal boundaries of sub-activities for discriminative clustering. Sener \etal~\cite{sener2015unsupervised}, represent every frame as a concatenated histogram of text and visual words, which are used as input to a probabilistic model. The applicability of these methods is limited because neither the existence of accompanying text, nor their proper alignment to the visual data can be taken for granted.

More recent works focus on developing weakly-supervised solutions, \ie where the orderings of the sub-activities are provided either only during training~\cite{huang2016connectionist, richard2017weakly} or testing as well~\cite{bojanowski2014weakly}. These methods try to align the frames to the given ordered sub-activities. Similar to us, the work of Bojanowski \etal~\cite{bojanowski2014weakly} includes a \emph{``background"} class. However, they assume that the background appears only once between every consecutive pair of sub-activities, while our model does not force any constraints on the occurrence of background. Others~\cite{huang2016connectionist, richard2017weakly} borrow temporal modelling methods from speech recognition such as connectionist temporal classification, RNNs and HMMs.

In the bigger scope of temporal sequences, several previous works have also addressed unsupervised segmentation~\cite{fox2014joint,Krueger15pami,zhou2013hierarchical}. Similar to us in spirit is the work of Fox \etal~\cite{fox2014joint}, which proposes a Bayesian nonparametric approach to model multiple sets of time series data concurrently. However, it has been applied only to motion capture data. Since skeleton poses are lower-dimensional and exhibit much less variance than video, it is unlikely for such a model to be directly applicable to video without a strong discriminative appearance model. To our knowledge, we are the first to tackle the problem of complex activity segmentation working solely with visual data without any supervision.

\section{The Generalized Mallows Model (GMM)}
~\label{sec:gmm} 
The GMM models distributions over orderings or permutations. In the standard Mallows model~\cite{mallows1957non}, the probability of observing some ordering $\pim$ is defined by a dispersion parameter $\rho$ and a canonical ordering $\sigmam$, 
\vspace{-0.25cm}
\begin{equation} 
 P_{\text{MM}}(\pim|\sigmam,\rho) = \frac{e^{-\rho\cdot d(\pim, \sigmam)}}{\psi(\rho)},
\end{equation}
\noindent where any distance metric for rankings or orderings can be used for $d(\cdot,\cdot)$. The extent to which the probability decreases as $\pim$ differs from $\sigmam$ is controlled by a dispersion parameter $\rho > 0$; $\psi(\rho)$ serves as a normalization constant.

The GMM, first introduced by Fligner and Verducci~\cite{fligner1986distance}, extends the standard Mallows model by introducing a set of dispersion parameters $\rhom = [\rho_1, ..., \rho_{K-1}]$, to allow individual parameterization of the $K$ elements in the ordering. The GMM represents permutations as a vector of inversion counts $\bm{v} = [v_1,..., v_{K-1}]$ with respect to an identity permutation $(1,...,K)$, where element $v_k$ corresponds to the total number of elements in $(k + 1, \ldots , K)$ that are ranked before $k$ in the ordering $\pim$\footnote{Only $K-1$ elements are needed since $v_K$ is 0 by definition as there cannot be any elements greater than $K$.}. If we assume that $\sigmam$ is the identity permutation, then a natural distance $d(\pim, \sigmam)$ can be defined as $\sum_k \rho_k v_k$, leading to 
\vspace{-0.45cm}
\begin{equation} \label{eq:computeV}
P_{\text{GMM}} (\bm{v} | \boldsymbol{\rho} ) = \frac{e^{-\sum_k \rho_k v_k}}{\psi_k(\boldsymbol{\rho})} = 
\prod_k^{K-1} \frac{e^{-\rho_k v_k} }{\psi_k(\rho_k)},
\end{equation} 
with $\psi_k(\rho_k)\!=\!\frac{1-e^{-(K-k+1) \rho_k}}{1-e^{-\rho_k}}$ as the normalization.
 
As the GMM is an exponential distribution, the natural prior for each element 
$\rho_k$ is the conjugate:
\vspace{-0.2cm}
\begin{equation}\label{eq:gmmRhoSampler}
P_{\text{GMM}_0}(\rho_k | v_{k,0}, \nu_0) \propto e^{-\rho_k v_{k,0} - log ( \psi_k (\rho_k))\nu_0 },
\end{equation} 
\noindent with hyper-parameters $v_{k,0}$ and $\nu_0$. Intuitively, the prior states that over $\nu_0$ previous trials, $\nu_0 \cdot v_{k,0}$ inversions will be observed~\cite{chen2009content}. For simplicity, we do not set multiple priors for each $k$ and use a common prior $\rho_0$ as per~\cite{chen2009content}, such that
\vspace{-0.2cm}
\begin{equation}
v_{k,0} = \frac{1}{e^{\rho_0 - 1}} - \frac{K-k + 1}{e^{(K-k+1) \rho_0} - 1}.
\end{equation}

\section{Proposed Model}
~\label{sec:method} 
Assume we are given a collection of $M$ videos, all of the same complex activity, and that each video is composed of an ordered sequence of multiple sub-activities. A single video $i$ with $J_i$ frames can be represented by a design matrix of features $\bm{F}_i \in \mathbb{R}^{J_i \times D}$, where $D$ is the feature dimension. We further define $\bm{F}$ as the concatenated design matrix of features from all $M$ videos and $\bm{F}_{\setminus i}$ as the features excluding video $i$. We first describe how we discriminatively learn the features $\bm{F}$ in Sec.~\ref{sec:visualfeat} before describing the standard temporal model in Sec.~\ref{sec:standardmodel} and the full model which models background frames in Sec.~\ref{sec:fullmodel}.

\subsection{Sub-Activity Visual Features}
~\label{sec:visualfeat}
Within a video collection of a complex activity there may be huge variations in visual appearance, even with state-of-the-art visual feature descriptors~\cite{sanchez2013image,tran2015learning,wang2013dense}. Suppose for frame $j$ of video $i$ we have video features $X_{ij}$ with dimensionality $V$. These features, if clustered naively, are most likely to group together according to video rather than sub-activity. To cluster the features more discriminantly, we learn a linear mapping of these features into a latent embedding space, \ie $\Phi_f(X_{ij}) : \mathbb{R}^V \rightarrow \mathbb{R}^E$. We also define in the latent space $K$ anchor points, with locations determined by a second mapping $\Phi_a(k) : \{1, \dots, K\} \rightarrow \mathbb{R}^E$. More specifically,
\vspace{-0.3cm}
\begin{align}
\Phi_f(X_{ij}) & = \bm{W}_f X_{ij}, \quad \bm{W}_f \in \mathbb{R}^{E \times V} \\
\Phi_a(k) & = \bm{W}_a(k), \quad \bm{W}_a \in \mathbb{R}^{E \times K}
\end{align} 
\noindent where $\bm{W}_f$ and $\bm{W}_a$ are the learned embedding weights and $E$ is the dimensionality of the joint latent space. Here, $\bm{W}_a(k)$ is the $k$-th column of $\bm{W}_a$, which corresponds to the location of anchor $k$ in the latent space. Together, $\bm{W}_f$ and $\bm{W}_a$ make up the parameter $\bm{W}$. We use the similarity of the video feature with respect to these anchor points as a visual feature descriptor, \ie
\vspace{-0.2cm}
\begin{equation}
\bm{F}_{ij} = {\bm{W}_a}^\intercal \bm{W}_f {X_{ij}}, 
\end{equation}
\noindent where $\bm{F}_{ij} = [f^1, ..., f^K]_{ij}$. Each element $f^k_{ij}$ is inversely proportional to the distance between $X_{ij}$ and anchor point $k$ in the latent space. By using $K$ anchor points, this implies that $D=K$.

Our objective in learning the embeddings is to cluster the video features discriminatively. We achieve this by encouraging the $X_{ij}$ belonging to the same sub-activity to cluster closely around a single anchor point while being far away from the other anchor points. If we assign each anchor point to a given sub-activity, then we can learn $\bm{W}$ by minimizing a pair-wise ranking loss $L$, where
\vspace{-0.25cm}
\begin{equation}\label{eq:rankingloss}
L = \sum_{i,j}^{M,J_i} \sum_{k=1,k \neq k^*}^K \max[0, f^{k}_{ij} - f^{k^*}_{ij} + \Delta] + \gamma ||\mathbf{W}||_2^2.
\vspace{-0.22cm}
\end{equation}
\noindent In this loss, $k^*$ is the anchor point associated with the true sub-activity label for $\bm{F}_{ij}$, $\Delta$ is a margin parameter and $\gamma$ is the regularization constant for the $l_2$ regularizer of $\bm{W}$. The loss in Eq.~\ref{eq:rankingloss} encourages the distance of $X_{ij}$ in the latent space to be closer to the anchor point $k^*$ associated with the true sub-activity than any other anchor point by a margin $\Delta$. 

The above formulation assumes that the right anchor point $k^*$, \ie the true sub-activity label, is known. This is not the case in an unsupervised scenario so we follow an iterative approach where we learn $\bm{W}$ at each iteration from an assumed sub-activity based on the segmentation of the previous iteration. More details are given in Sec.~\ref{sec:learn}.
 
\subsection{Standard Temporal Model}
~\label{sec:standardmodel}
Given a collection of $M$ videos of the same complex activity, we would like to infer the sub-activity assignments $\bm{z} = \{\bm{z}_{i}\}, i\in\{1,\ldots,M\}$. For video $i$, $\bm{z}_i = \{z_{ij}\}, j\in\{1,\ldots,J_i\}$, $z_{ij} \in \{1,\ldots,K\}$ can be assigned to one of $K$ possible sub-activities\footnote{For convenience, we overload the use of $K$ for both the number of elements in the ordering for the GMM as well as the number of sub-activities, as the two are equal when applying the GMM.}. 
We introduce $\bm{a}_i$, a bag of sub-activity labels for video $i$, \ie the collection of elements in $\bm{z}_i$ but without consideration for the temporal frame ordering. The ordering is then described by $\pim_i$. $\bm{a}_i$ is expressed as a vector of counts of the $K$ possible sub-activities, while $\pim_i$ is expressed as an ordered list. Together, $\bm{a}_i$ and $\pim_i$ determine the sub-activity label assignments $\bm{z}_i$ to the frames of video $i$. $({\bm{a},\pim})$ are redundant to $\bm{z}$; the extra set of variables gives us the flexibility to separately model the sub-activities' visual appearance (based on $\bm{a}$) from the temporal ordering (based on $\pim$). We model $\bm{a}$ as a multinomial, with parameter $\thetam$ and a Dirichlet prior with hyperparameter $\theta_0$. For the ordering $\pim$, we use a GMM with the exponential prior from Eq.~\ref{eq:gmmRhoSampler} and hyperparameters $\rho_0$ and $\nu_0$. 
The joint distribution of the model factorizes as follows:
\vspace{-0.2cm}
\begin{equation}\label{eq:fulljoint}
\begin{split}
P(\bm{z}, \thetam, & \rhom, \bm{F} | \theta_0, \rho_0, \nu_0) \\ 
= &P(\bm{F}| \bm{z}) P(\bm{a} |\thetam) P(\pim | \rhom) P(\thetam | \theta_0) P(\rhom | \rho_0, \nu_0) \\
 = & \Big [ \prod_{i,j=1}^{M,J_i} %\prod_{j=1}^{J_i} 
P(\bm{F}_{ij}|z_{ij} ) \Big ] 
 \Big [ \prod_{i=1}^M P(\bm{a}_i |\thetam) P(\pim_i | \rhom) \Big ] \\ 
 &\Big [ \prod_{k=1}^K P(\theta_k | \theta_0) \Big ] 
\Big [ \prod_{k=1}^{K-1} P(\rho_k | \rho_0, \nu_0) \Big ],
\vspace{-0.5cm}
\end{split}
\end{equation}
based on the assumption that each frame of each video as well as each video are all independent observations.
 
\begin{figure}[t!]
 \centering
	\begin{minipage}{1.0\linewidth} 
	\begin{subfigure}[b]{0.45\textwidth}
	\resizebox{\linewidth}{!}{
	\scalebox{0.1}{

 	\begin{tikzpicture}[x=1.7cm,y=1.6cm] 
 		\node[obs]                                             (F)       {$\bm{F}$} ; 
 		\node[latent, above=of F, yshift=-1.3cm]               (z)       {$\bm{z}$} ; 
 		\node[latent, left=of z, yshift=0.9cm, xshift=1.3cm]   (a)       {$\bm{a}$} ;  
 		\node[latent, above=of a, yshift=-0.3cm]               (theta)   {$\thetam$} ; 
 		\node[disc, above=of theta, yshift=-0.6cm]             (ztheta)  {$\theta_0$} ; 
 		\node[latent, right=of z, yshift=0.9cm, xshift=-1.3cm] (pi)      {$\pim$} ;  
 		\node[latent, above=of pi, yshift=-1.4cm]              (v)       {$\bm{v}$} ; 
 		\node[latent, above=of pi, yshift=-0.3cm]              (rho)     {$\rhom$} ; 
 		\node[disc, above=of rho, yshift=-0.6cm]               (rhozviz) {$\rho_0,\nu_0$} ; 
 		\edge {theta} {a} ; 
 		\edge {ztheta} {theta} ;  
 		\edge {z} {F} ;  
     	\draw[->,dashed](v)--(pi) ;  
 		\edge {rho} {v} ;  
 		\edge {rhozviz} {rho} ; 
 		\draw[->,dashed](a)--(z) ; 
 		\draw[->,dashed](pi)--(z) ; 
 		\plate {plate1} { (F) (z) } {$J_i$}; 
 		\plate {} { (plate1) (a) (pi) (v) } {$M$} ;   
	\end{tikzpicture} 
	
	} }
	\caption{\footnotesize{Standard model}}
	\label{fig:subfigModel}
	\end{subfigure} 
	\begin{subfigure}[b]{0.45\textwidth}
	\resizebox{\linewidth}{!}{
	\scalebox{0.1}{

 	\begin{tikzpicture}[x=1.7cm,y=1.6cm] 
 		\node[obs]                                             (F)         {$\bm{F}$} ; 
 		\node[latent, above=of F, yshift=-1.3cm]               (z)         {$\bm{z}$} ; 
 		\node[latent, left=of z, yshift=0.9cm, xshift=1.3cm]   (a)         {$\bm{a}$} ;  
 		\node[latent, above=of a, yshift=-0.3cm]               (theta)     {$\thetam$} ;  
 		\node[disc, above=of theta, yshift=-0.6cm]             (ztheta)    {$\theta_0$} ; 
 		\node[latent, right=of z, yshift=0.9cm, xshift=-1.3cm ]              (pi)        {$\pim$} ; 
 		\node[latent, above=of pi, yshift=-1.4cm]              (v)         {$\bm{v}$} ; 
 		\node[latent, above=of pi, yshift=-0.3cm]              (rho)       {$\rhom$} ; 
 		\node[disc, above=of rho, yshift=-0.6cm]               (rhozviz)   {$\rho_0,\nu_0$} ; 
 		\node[latent, above=of z, yshift=-1.4cm ] (b)         {$\bm{b}$} ; 
 		\node[latent, above=of b, yshift=-0.3cm]               (lambda)    {$\lambda$} ;  
 		\node[disc, above=of lambda, yshift=-0.6cm]            (alphabeta) {$\alpha,\beta$} ; 
 		\edge {theta} {a} ; 
 		\edge {ztheta} {theta} ;  
 		\edge {z} {F} ;  
 		\draw[->,dashed](v)--(pi) ;  
 		\edge {rho} {v} ;  
 		\edge {rhozviz} {rho} ; 
 		\edge {lambda} {b} ; 
 		\edge {alphabeta} {lambda} ; 
 		\draw[->,dashed](a)--(z) ; 
 		\draw[->,dashed](pi)--(z) ; 
 		\draw[->,dashed](b)--(z) ; 
 		\plate {plate1} { (F) (z) (b) } {$J_i$}; 
 		\plate {} { (plate1) (a) (pi) (v) } {$M$} ;  
	\end{tikzpicture} 
	
	} }
	\caption{\footnotesize{Full model with background}}
	\label{fig:subfigModelBck}
	\end{subfigure} 
	\end{minipage} 
\vspace{-0.3cm}
 \caption{Plate diagrams of our models. Shaded nodes: observed variables, rectangles: fixed hyper-parameters, dashed arrows: deterministically constructed variables.
 }
 \vspace{-0.3cm}
\end{figure}
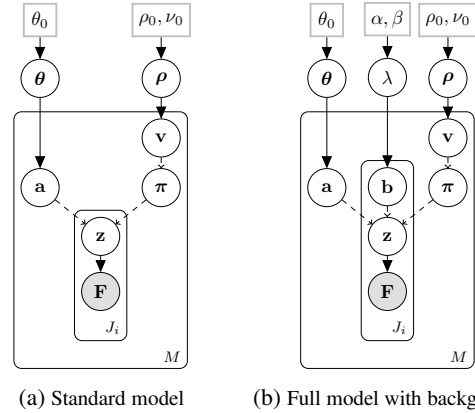

We show a diagram of the model in Fig.~\ref{fig:subfigModel}. When using the GMM, performing MLE to find a consensus or canonical ordering over a set of observed orderings is an NP hard problem, though several approximations have been proposed. Our case is the reverse, in which we assume that a canonical ordering is already given and we would like to find a (latent) set of orderings. Our interest is to infer the posterior $P(\bm{z}, \rhom |\bm{F}, \theta_0, \rho_0, \nu_0 )$ for the entire video corpus. Directly working with this posterior is intractable, so we make MCMC sampling-based approximations. Specifically, we use slice sampling for $\boldsymbol{\rho}$ and collapsed Gibbs sampling~\cite{griffiths2004finding} for $\bm{z}$. Since $\bm{z}$ is fully specified by $\bm{a}$ and $\pim$, it is equivalent to sample $\bm{a}$ and $\pim$. Before elaborating on the sampling equations, we first detail how we model the video likelihood $P(\bm{F}_{i} | \bm{z}_i)$.

\vspace{-0.3cm}
\paragraph{Video likelihood $P(\bm{F}_{i} | \bm{z}_i)$} \hspace{-0.2cm} can be broken down into the product of frame likelihoods, since each frame is conditionally independent given the frame's sub-activity, \ie
\vspace{-0.35cm}
\begin{equation}\label{eq:computeFeatureLikelihood} 
P(\bm{F}_i | \bm{z}_i, \bm{F}_{\setminus i}, \bm{z}_{\setminus i}) = \prod_{j=1}^{J_i} P(\bm{F}_{ij} | z_{ij}, \bm{F}_{\setminus i}, \bm{z}_{\setminus i} ). 
\vspace{-0.3cm}
\end{equation} 
\noindent Since our temporal model is generative, we need to make some assumptions about the generating process behind the video features. We directly model the frame likelihoods and use $K$ mixtures of Gaussians, one for each sub-activity $k$. Each mixture has $Q$ components with weights $\omega_k$, means $\boldsymbol{\mu}_k$ and covariances $\bm{\Sigma}_k$, with likelihood scores for each mixture selected according to the assignments $z_{ij}$:
\vspace{-0.35cm}
\begin{equation}~\label{eq:framelikelihood}
 P(\bm{F}_{ij} | z_{ij} = k, \bm{F}_{\setminus i}, \bm{z}_{\setminus i}) \sim \sum_{q=1}^Q\omega^q_k \cdot \mathcal{N}(\boldsymbol{\mu}^q_{k}, \bm{\Sigma}_k).
\end{equation}
 
\vspace{-0.7cm}
\paragraph{Sampling sub-activity $\bm{a}_{i}$} \hspace{-0.3cm} is done with collapsed Gibbs sampling. Recall that $\bm{a}$ is modelled as a multinomial with $K$ outcomes parameterized by $\thetam$. We sample $a_{ij}$, the $j\text{-th}$ frame for video $i$, from the posterior conditioned on all other variables. Without the redundant terms, this posterior is expressed as
\vspace{-0.3cm}
\begin{equation}\label{eq:subActProb} 
\resizebox{0.88\hsize}{!}{ 
$P( a_{ij}=k | \dots ) \propto P(a_{ij} =k|\bm{a}_{\setminus ij}, \theta_0 ) \cdot P(\bm{F}_i | \bm{z}_i, \bm{F}_{\setminus i}, \bm{z}_{\setminus i}),$
}
\end{equation}
\noindent where the second term is the video likelihood from Eq.~\ref{eq:computeFeatureLikelihood}. The first term is a prior over the sub-activities, and can be estimated by integrating over $\thetam$. The integration is done via the collapsed Gibbs sampling, and, as we assumed that 
$\thetam \sim \text{Dirichlet}(\theta_0)$, this results in 
\vspace{-0.4cm}
\begin{equation} \label{eq:computeBags} 
P(a_{ij}\!= k | a_{\setminus ij}, \theta_0 ) = 
 \frac{N_k + \theta_0 }{ \sum_{k=1}^K N_k + K \theta_0 },
\end{equation}
where $N_k$ is the total number of times the sub-activity $k$ is observed in the all sequences and $\sum_{k=1}^K N_k$ is the total number of sub-activity assignments. 

Note that sampling ${a}_{ij}$ does not correspond to the sub-activity assignment to the $j\text{-th}$ frame. The assignment is given by ${z}_{ij}$, which can only be computed after sampling $a_{ij}$ for all $J_i$ frames of video $i$ and then re-ordering the bag of frames according to $\pim_i$.

\vspace{-0.4cm}
\paragraph{Sampling ordering $\pim_i$} \hspace{-0.2cm} is done via regular Gibbs sampling. Recall that the ordering follows a GMM as described in Sec.~\ref{sec:gmm} and is parameterized for elements in the ordering individually via inversion count vector $\bm{v}_i$. As such, we sample a value for each position in the inversion count vector from $k=1$ to $K-1$ independently according to
\vspace{-0.2cm} 
\begin{equation} \label{eq:temporalModel} 
\resizebox{.88\hsize}{!}{$
P(v_{ik} = c | \bm{z}, \rhom, \bm{F}) \propto P( v_{ik} = c | \rho_k), \cdot P(\bm{F}_i | \bm{z}_i, \bm{F}_{\setminus i}, \bm{z}_{\setminus i}) $,} 
\end{equation}
where $c$ indicates the inversion count assignment to $v_{ik}$. Again, the second term is the video likelihood from Eq.~\ref{eq:computeFeatureLikelihood}, while the first term corresponds to $P_{\text{GMM}}( v_{ik} = c; \rho_k) $, and is computed according to Eq.~\ref{eq:computeV}. We estimate the probability of every possible value of $v_{ik}$, which ranges from 0 to $K-k$, and sample a new inversion count value $c$ based on these probabilities.

\vspace{-0.4cm}
\paragraph{Sampling GMM dispersion parameter $\bm{\rho_k}$:} This is done for each sub-activity $k = 1$ to $K-1$ independently. We draw $\rho_k$ using slice sampling~\cite{mackay2003information} from the conjugate prior distribution $P_{\text{GMM}_0}$ according to Eq.~\ref{eq:gmmRhoSampler}. 

\subsection{Background Modeling}~\label{sec:fullmodel}
To consider background, we extend the label assignment vector $\bm{z}$ with a binary indicator variable $b_{ij} \in \{0,1\}$ for each frame. The indicator $b_{ij}$ follows a Bernoulli variable parameterized by $\lambda$, with a beta prior, \ie $\lambda \sim \ \text{Beta} ( \alpha, \beta ) $. 
In this setting, $\bm{z}_i$ is determined by the bag of sub-activities $\bm{a}_i$, the ordering $\pim_i$, and background vector $\bm{b}_i = \{b_{ij}\}$, where $\bm{b}_i$ indicates the frames to be excluded from sub-activity consideration. For example, for video $i$, given $\bm{a}_i\!=\![6 \,\, 3 \,\, 5]$, $\pim_i\!=\![2 \,\, 3 \,\, 1]$ and $\bm{b}_i\!=\![1 1 1 0 0 1 1 1 0 0 1 1 0 0 0 1 1 1 1 0 0 1 1 ]$, the sub-activity assignment is $\bm{z}_i\!=\![2 2 2 0 0 3 3 3 0 0 3 3 0 0 0 1 1 1 1 0 0 1 1 ]$. 

We show a diagram of the model in Fig.~\ref{fig:subfigModelBck}. The joint distribution of the model can be expressed as 
\vspace{-0.2cm}
\begin{equation}~\label{eq:posteriorfull}
\begin{split}
P(\bm{z}, & \thetam, \rhom, \bm{F} | \theta_0, \alpha, \beta, \rho_0, \nu_0, ) = P(\bm{a} |\thetam, \theta_0 ) \\ \cdot & P( \pim | \rhom, \rho_0, \nu_0 ) 
\cdot P(\bm{b} | \lambda, \alpha, \beta) \cdot P(\bm{F} | \bm{a}, \pim, \bm{b} ).
\end{split} 
\end{equation}

\noindent Drawing samples from this full model requires a small modification to the sub-activity sampling $\bm{a}_i$. More specifically, we need a blocked collapsed Gibbs sampler that samples ${a}_{ij}$ and $b_{ij}$ jointly while integrating over $\thetam$ and $\lambda$.

\vspace{-0.4cm}
\paragraph{Sampling background $\bm{b}_{i}$} \hspace{-0.3cm} is done from the joint conditional 
\vspace{-0.2cm}
\begin{equation}~\label{eq:samplebk1}
\resizebox{.87\hsize}{!}{$
\!\!\! P(b_{ij}, a_{ij} | \dots ) \propto P(b_{ij} | \alpha, \beta) \cdot P(a_{ij} |\bm{a}_{\setminus ij}, \theta_0 ) \cdot P(\bm{F}_i | \bm{z}_i, \bm{F}_{\setminus i}, \bm{z}_{\setminus i}).$}
\end{equation}
This is equivalent to the following for a sub-activity frame:
\vspace{-0.2cm}
{\small
\begin{equation}~\label{eq:bk1}
\begin{split}
\!\!\!\!\!P(& b_{ij}\!=\!0, a_{ij}\!=\!k |\dots ) \propto \frac{N_f + \alpha }{N_f + N_b +\alpha +\beta }\\
& \cdot \frac{N_k + \theta_0 }{ \sum_{k=1}^K N_k + K \theta_0 } \cdot P(\bm{F}_i | b_{ij}\!=\!0, a_{ij}\!=\!k, \bm{F}_{\setminus i},\bm{z}_{\setminus ij}) , \end{split}
\end{equation}
}
where $N_f$ and $N_b$ are the total number of sub-activity frames and background frames in the corpus respectively. For a background frame, the joint conditional is equal to
\vspace{-0.2cm}
\begin{equation}~\label{eq:bk2}
\resizebox{.87\hsize}{!}{$
\!\!\!\! P(b_{ij}\!=\!1, a_{ij}|\dots) \propto \frac{N_b + \alpha }{N_f + N_b +\alpha +\beta } \cdot P(\bm{F}_i | b_{ij}\!=\!1, a_{ij},\bm{F}_{\setminus i},\bm{z}_{\setminus ij} ). $
}
\end{equation}
The video likelihood in Eqs.~\ref{eq:bk1} and~\ref{eq:bk2} are computed in a similar way as defined in Eqs.~\ref{eq:computeFeatureLikelihood} and~\ref{eq:framelikelihood}, with the exception that we now iterate over the joint states of background and sub-activity labels for the frame likelihoods. Note that this only adds one extra probability in being computed, \ie $b\!=\!1$, since the state of $a_{ij}$ is then irrelevant. The rest of the Gibbs sampling remains the same. 

\begin{figure*}[htb]
	\centering 
		\includegraphics[height=4.1cm]{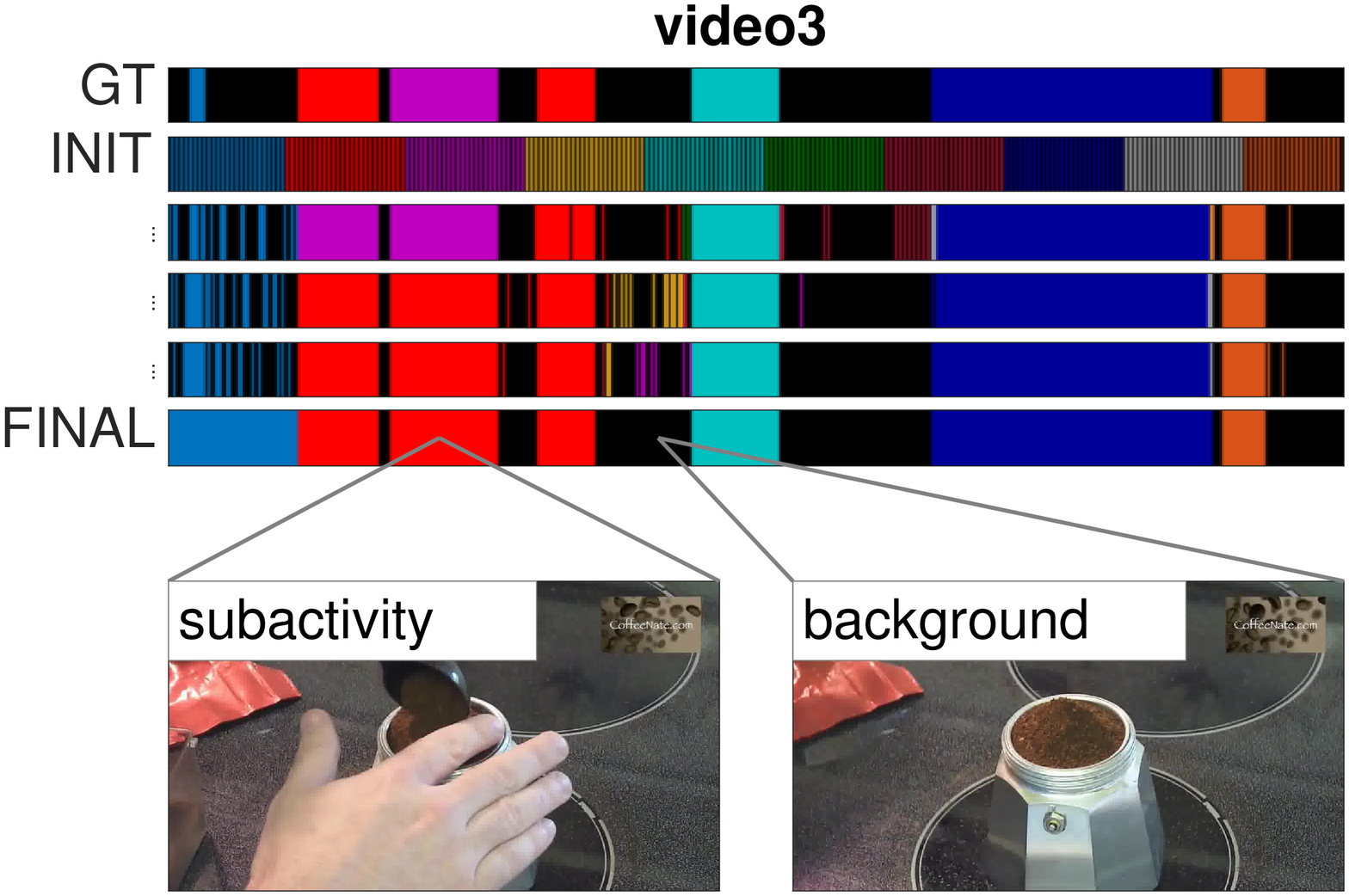} 
		\includegraphics[height=4.1cm]{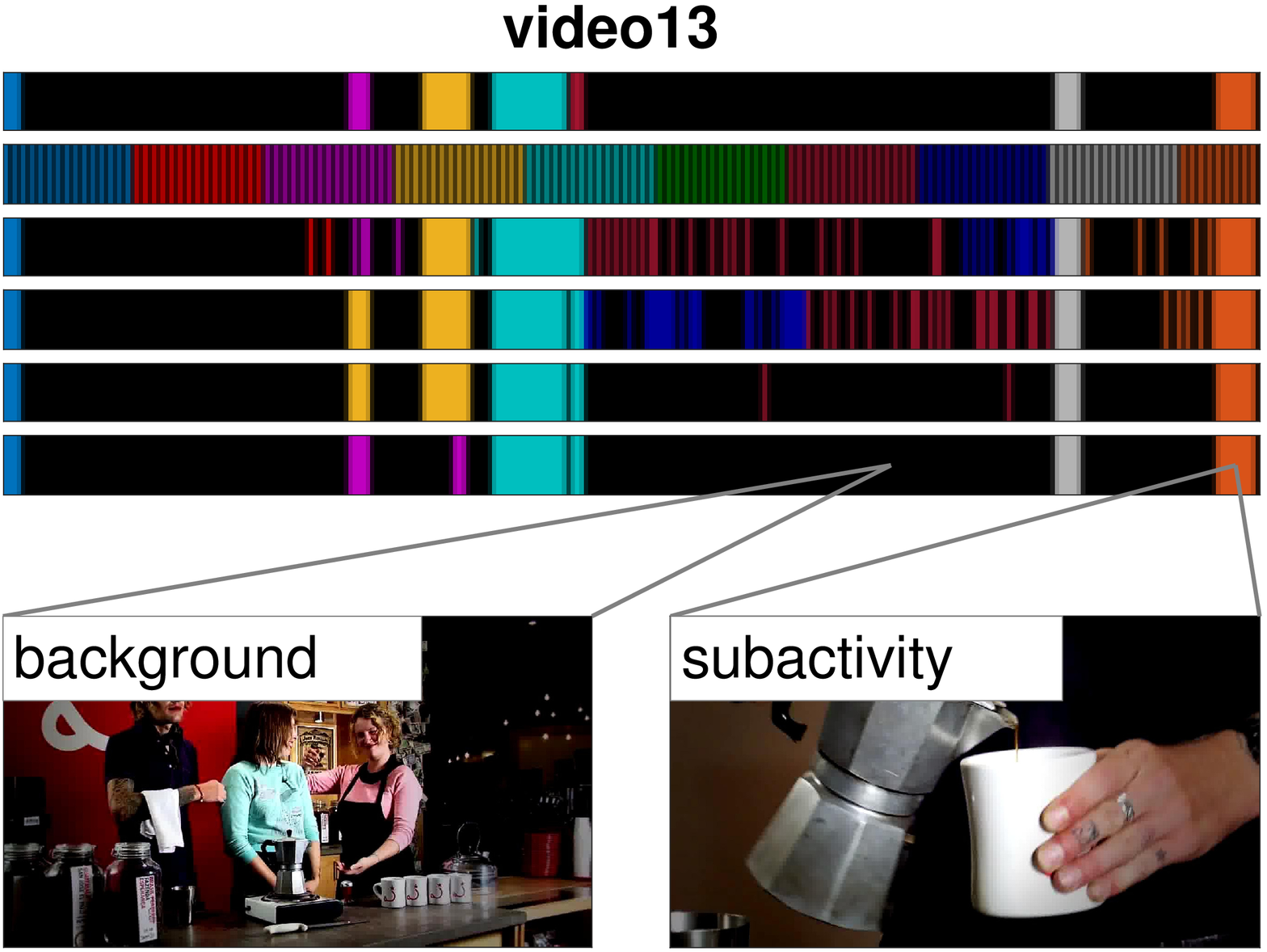}
		\includegraphics[height=4.1cm]{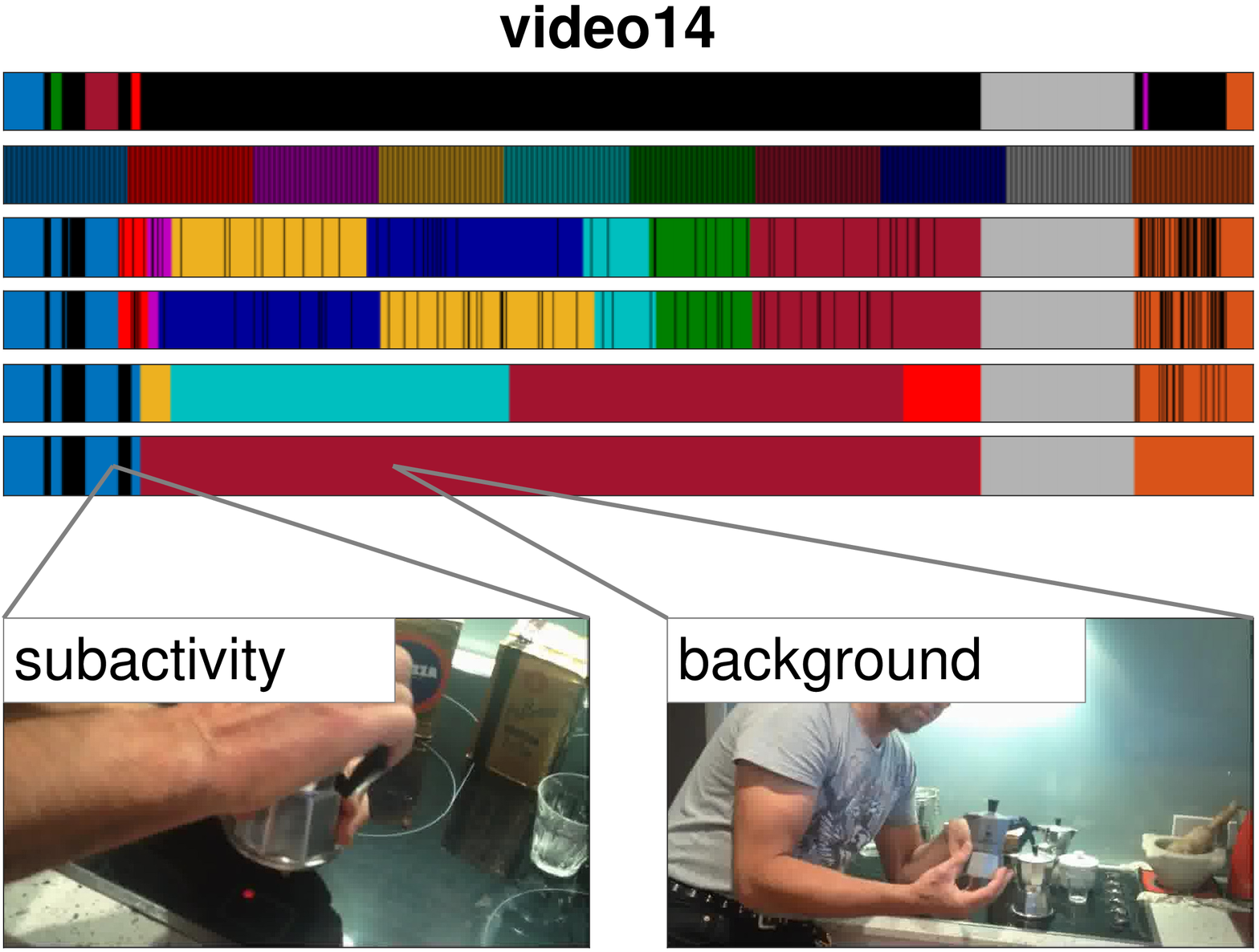} 
\vspace{-0.2cm}
	\caption{
	Segmentation outputs on three \emph{`making coffee'} examples from Inria Instructional Videos Dataset~\cite{alayrac2016unsupervised}. Colors indicate different sub-activities, black the background frames. Since our algorithm is fully unsupervised, we established one-to-one color mappings between the ground truth and our outputs for visualization purposes. The first row (GT) is the ground truth; the remaining rows show the progression from the initialization (INIT) over some iterations to the (FINAL) segmentation. Our method performs well when the appearance of the sub-activities is discriminative,~\eg for video 3, occurrence of a hand during a sub-activity vs. none during the background frames, or people talking for video 13. We fail in detecting background when there are also interactions with objects of interest,~\eg in video 14. Our model does not enforce continuity over the background frames and may result in fragmentation, but as shown, with good appearance modelling, the background clusters naturally. Furthermore, the final segmentations may contain a different number of sub-activities while still maintaining a global order, \eg the orange sub-activity tends to appear last and follows the grey one.} 
\vspace{-0.4cm}
	\label{fig:segmentation_visual} 
\end{figure*}

\subsection{Inference Procedure} 
~\label{sec:learn}
Our model's inputs are the frames $\bm{X}$, the number of sub-activities $K$ and the number of Gaussian mixtures $Q$. We iterate between solving for $\bm{F}$ and sampling $\bm{z}$ and $\rhom$ from the posterior $P(\bm{z}, \rhom |\bm{F}, \theta_0, \alpha, \beta, \rho_0, \nu_0)$. To initialize $\bm{z}_i$ for each video $i$, the sub-activity counts $\bm{a}_i$ are split uniformly over $K$ sub-activities; $\pim_i$ is set to the canonical ordering; $\bm{b}_i$ is set with every other frame being background (see Fig.~\ref{fig:overviewFigure}). Using the current assignments $\bm{z}$, we first learn $\bm{W}$ of the latent embeddings to solve for $\bm{F}$ and then for each sub-activity $k$, the Gaussian mixture components $\{\bm{\omega}_k, \boldsymbol{\mu}_k, \bm{\Sigma}_k\}$. 
For each video $i$, we then proceed to re-sample $\{\bm{a}_i, \bm{b}_i\}$, $\pim_i$, in that order, using Gibbs sampling to construct $\bm{z}_i$. After repeating for each video, we can then re-sample the dispersion parameter $\rhom$. From the new $\bm{z}$ and $\rhom$, we then repeat. This process is summarized in the algorithm in Fig.~\ref{fig:overviewAlgo}.
 
To optimize Eq.~\ref{eq:rankingloss} for learning $\bm{W}$, we use Stochastic Gradient Descent (SGD) with mini-batches of 200 and momentum of 0.9. We set the hyper-parameters $\rho_0=1$, $\alpha = 0.2$, $\beta = 0.2$, $\nu_0= 0.1$, $\theta_0 = 0.1$. 
 
\section{Experimentation}
\label{experiments}

\begin{figure*}[htb]
	\centering
	\begin{subfigure}[b]{0.24\textwidth}
		\includegraphics[width=\columnwidth]{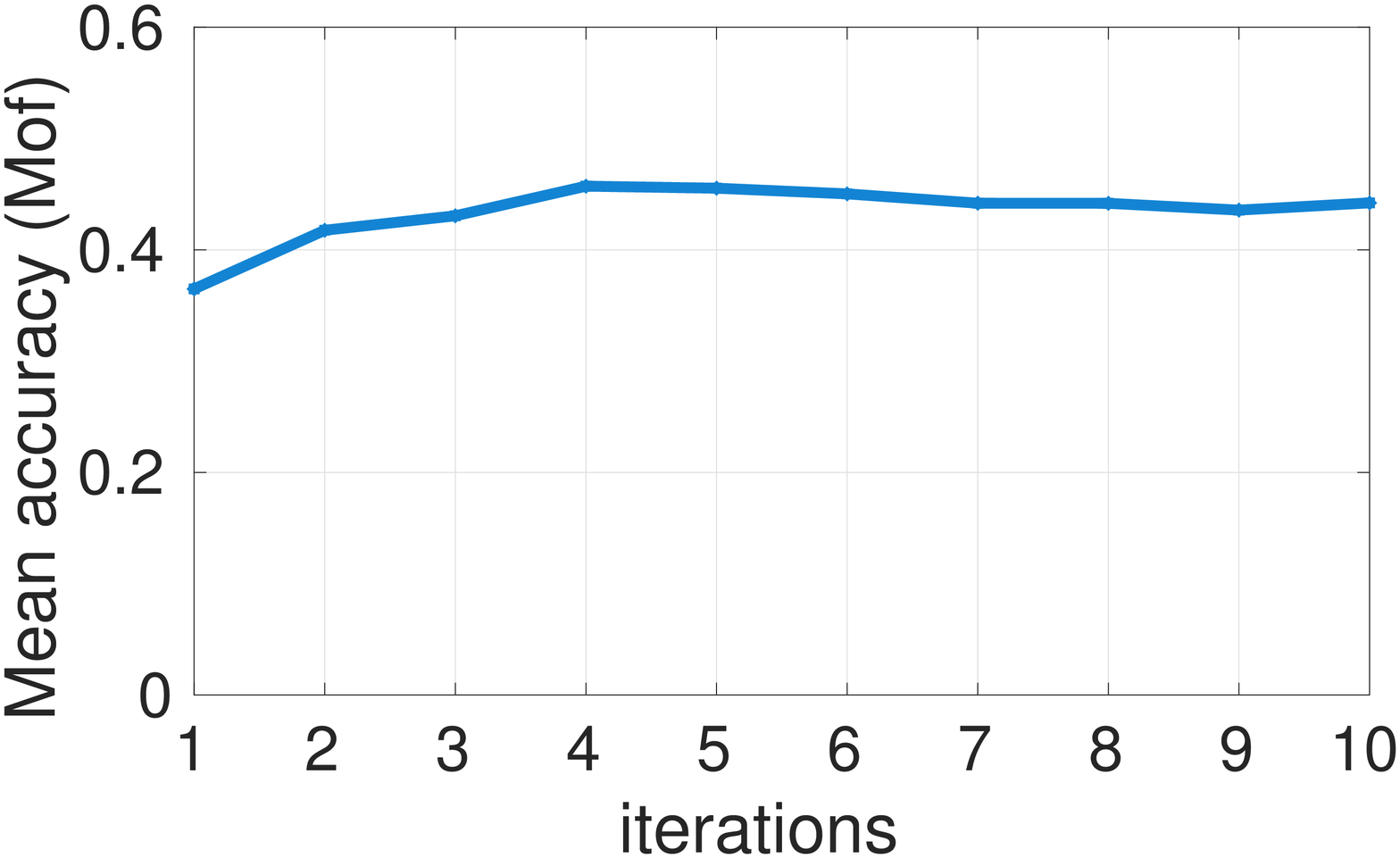}
		\caption{Convergence}
	\end{subfigure}
 \begin{subfigure}[b]{0.23\textwidth}
	 \includegraphics[width=\columnwidth]{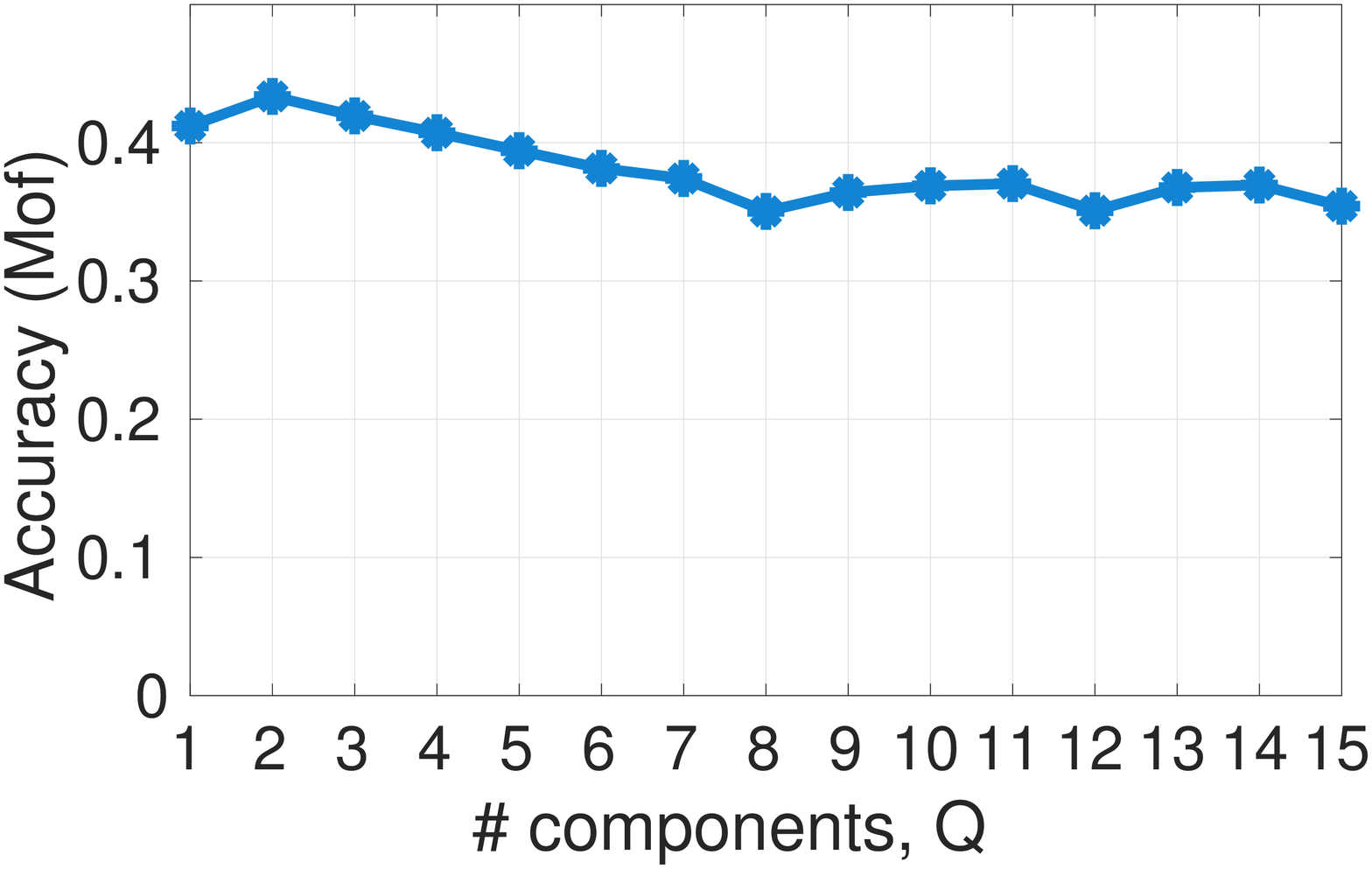}
	  \caption{vs. \# mixture components}
	 \end{subfigure}
 \begin{subfigure}[b]{0.24\textwidth}
		\includegraphics[width=\columnwidth]{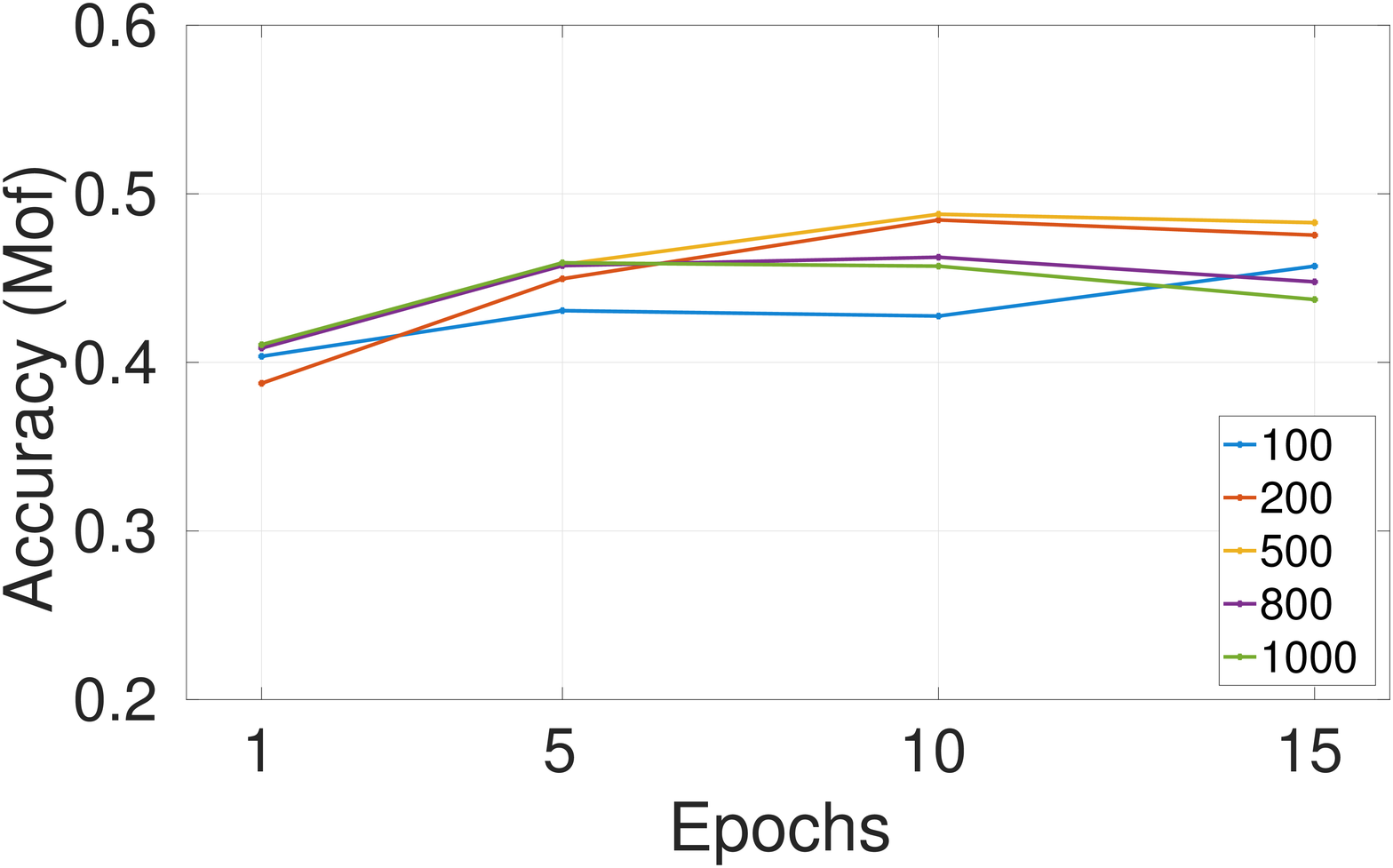}
		\caption{Dimensionality}
	\end{subfigure}	
\vspace{-0.3cm}
	\caption{Influence of our model's parameters are tested on the Instructional Videos Dataset~\cite{alayrac2016unsupervised} without background frames. We set $K$ to the ground truth sub-activity number of all five activities. Our method's performance over iterations is shown in (a), using different numbers of Gaussian mixture components in (b) and dimensionality of embedding space in (c).} 
\label{fig:inria_parameter_influence}
\vspace{-0.4cm}
\end{figure*}
 
\subsection{Datasets \& Evaluation Metrics}
We analyze our model's performance on two challenging datasets, Breakfast Actions~\cite{kuehne2014language} and Inria Instructional Videos~\cite{alayrac2016unsupervised}. Breakfast Actions has 1,712 videos of 52 parti\-cipants performing 10 breakfast preparation activities. There are 48 sub-activities, and videos vary according to the participants' preference of preparation style and orderings. We use the visual features from~\cite{kuehne2016end} based on improved dense trajectories~\cite{wang2013action}. This dataset has no background. 

Inria Instructional Videos contains 150 narrated videos of 5 complex activities collected from YouTube. The videos are on average 2 minutes long with 47 sub-activities. We use the visual features provided by~\cite{alayrac2016unsupervised}: improved dense trajectories and VGG-16~\cite{Simonyan14c} conv5 layer responses taken over multiple windows per frame. The trajectory and CNN features are each encoded with bag-of-words and concatenated for each frame. The videos are labelled, including the background, \ie frames in which the sub-activity is not visually discernible, usually when the person stops to explain past, current or upcoming steps. As such, the sub-activities are separated by hundreds of background frames (73\% of all frames). We evaluate our standard model without background modelling by removing these frames from the sequence as well as our full model on the original sequences.

To evaluate our segmentations in the fully unsupervised setting, we need one-to-one mappings between the segment and ground truth labels. In line with~\cite{alayrac2016unsupervised,sener2015unsupervised}, we use the Hungarian method to find the mapping that maximizes the evaluation scores and then evaluate with three metrics: The mean over frames (Mof) evaluates temporal localization of sub-activities and indicates the percentage of frames correctly labelled. The Jaccard index, computed as intersection over detections, as well as the $\text{F}_1$ score quantify differences between ground truth and predicted segmentations. With all three measures, higher values indicate better performance.

We also show a partly supervised baseline in which we use ground truth sub-activity labels for learning $\bm{F}$ but learn the temporal alignments unsupervised. This can be thought of as an upper bound on performance for our fully unsupervised version, in which we iteratively learn the temporal alignment and discover the visual appearance of the sub-activities. We refer to these to as \emph{``ours GT''} and \emph{``ours iterated''} respectively in the experimental results. 

\subsection{Sub-Activity Visual Appearance Modelling}
By projecting the frames' visual features and the sub-activity labels into a joint feature space, we learn a visual appearance model for the sub-activities. We first consider the standard model on Inria Instructional Videos with the background frames removed. The plot in Fig.~\ref{fig:inria_parameter_influence}(a) tells us that the appearance model can be learned successfully in an iterative fashion and begins to stabilize after approximately 5 iterations between learning the sub-activity appearance and the GMM. Our model's performance depending on the the number of Gaussian mixture components $Q$ is shown in Fig.~\ref{fig:inria_parameter_influence}(b). The resulting sub-activity representations are very low-dimensional and highly separable so that we achieve higher Mof with a few number of components. We use $Q=3$ mixture components for our iterative and $Q=1$ for the ground truth experiments. In Fig.~\ref{fig:inria_parameter_influence}(c), we use our iterated method to show the Mof for different values of $E$, our embedding dimensionality, over the training epochs. We find only small differences in Mof for different $E$ values. We fix the embedding size $E=200$ with 12 epochs of training and 5 iterations of sub-activity representation and GMM learning for subsequent experiments on both datasets. The run time of a single iteration of our algorithm is proportional to the number of frames $J_i$ in each video and the assumed number of sub-actvities $K$. On a computer with an Intel Core i7 3.30 GHz CPU, our model, for a single iteration, takes approximately $115$ seconds ($109$ for learning the sub-activity appearance model and $6$ seconds for estimating the temporal structure). 
 
\begin{figure}[h]
	\centering
	\begin{subfigure}[b]{0.47\columnwidth}
		\includegraphics[width=\columnwidth]{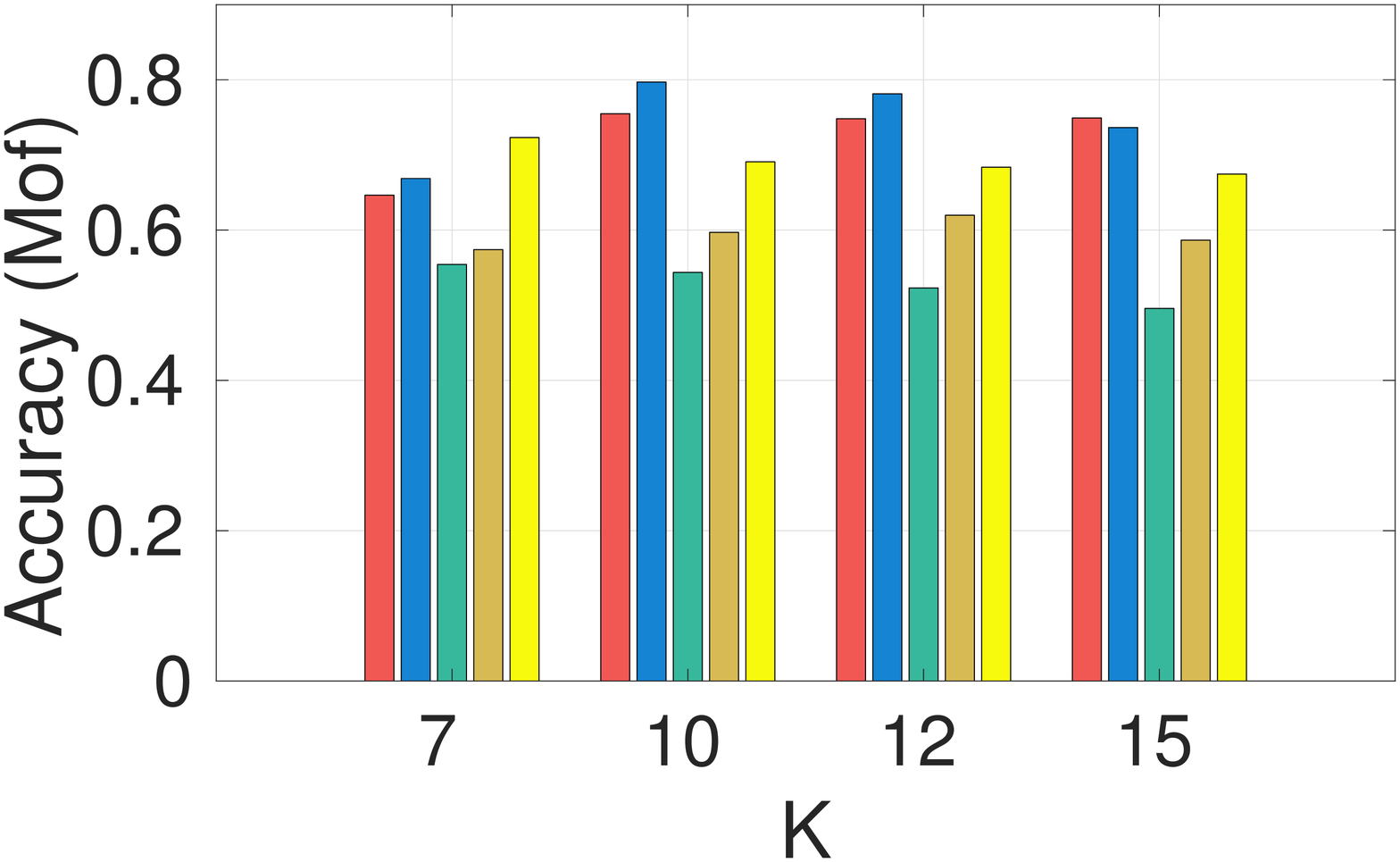}
		\vspace{-0.6cm}
		\caption{\textit{ours GT}}	
		\label{fig:noBck_gt_mof}
	\end{subfigure}~
	\begin{subfigure}[b]{0.47\columnwidth}
		\includegraphics[width=\columnwidth]{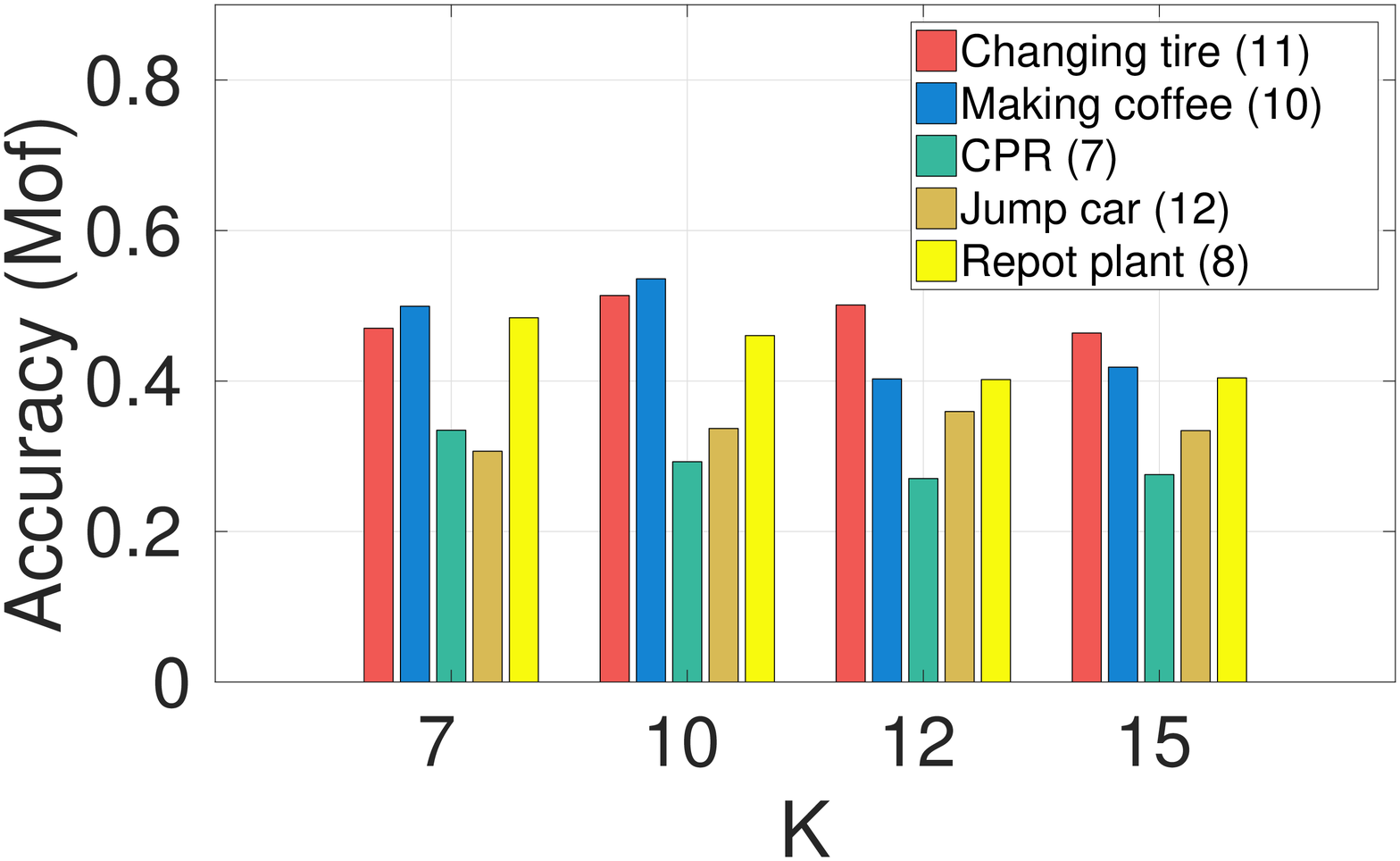}
		\vspace{-0.6cm}
		\caption{\textit{ours iterated}}
		\label{fig:noBck_iter_mof}
	\end{subfigure}~
\vspace{-0.2cm}
	\caption{Results (Mof) on Instructional Videos~\cite{alayrac2016unsupervised} without background frames with varying $K$. The legend gives the ground truth $K$ for each subactivity in braces. }
	\label{fig:inria_without_background}	
	\vspace{-0.3cm}
\end{figure}

\subsection{Temporal Structure Modelling}
The GMM models temporal ordering -- without it, one can only classify each frame's sub-activity label based on the visual appearance. Even if these appearance models are trained on ground truth, the segmentation results would be very poor. On Inria Instructional Videos without background, the average MoF over actions is 0.322 without versus 0.692 with the GMM (see Fig.~\ref{fig:inria_without_background}).

\begin{figure*}[t!]
	\centering 
	\includegraphics[height=2.8cm]{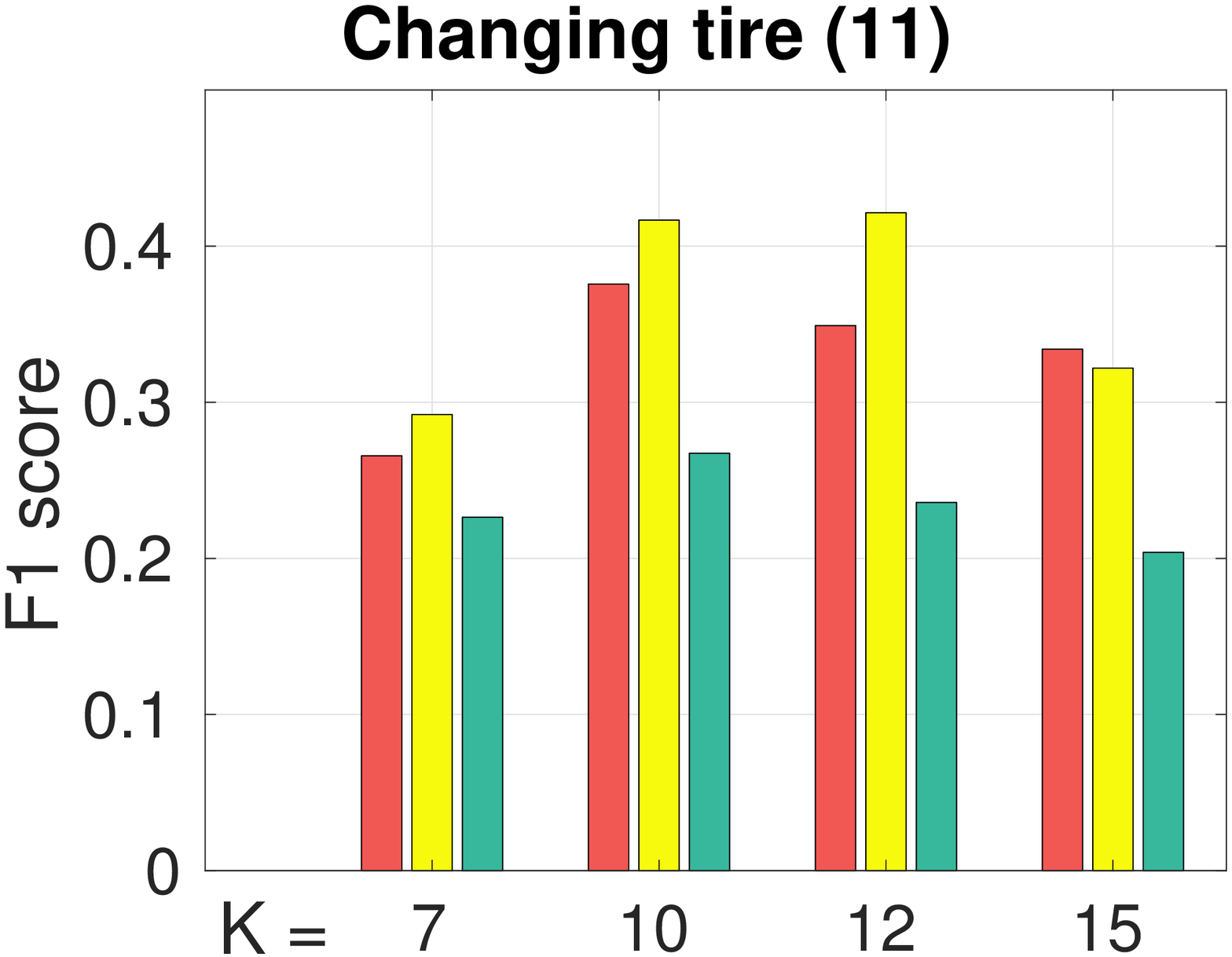}
	\includegraphics[height=2.8cm]{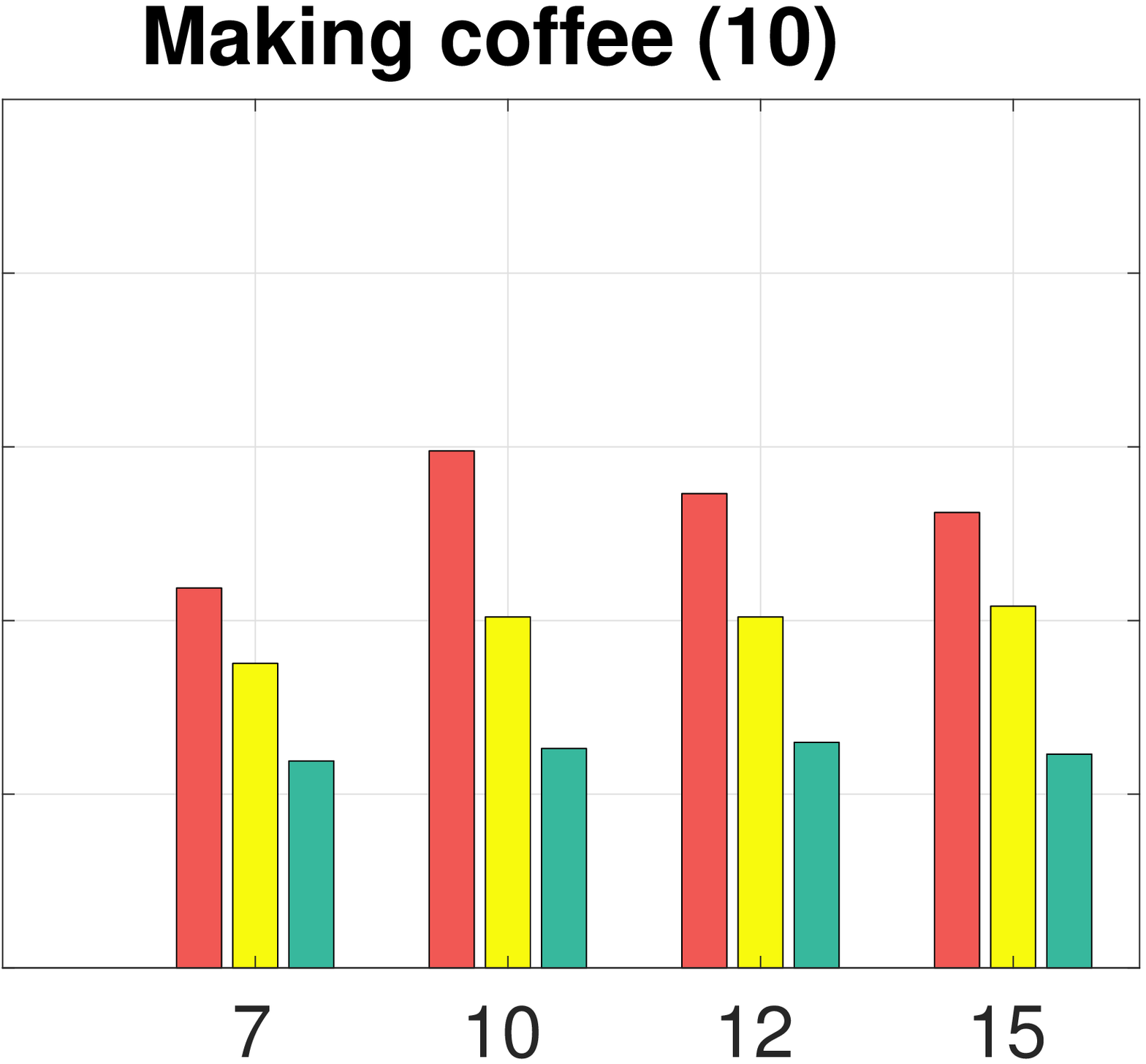}
	\includegraphics[height=2.8cm]{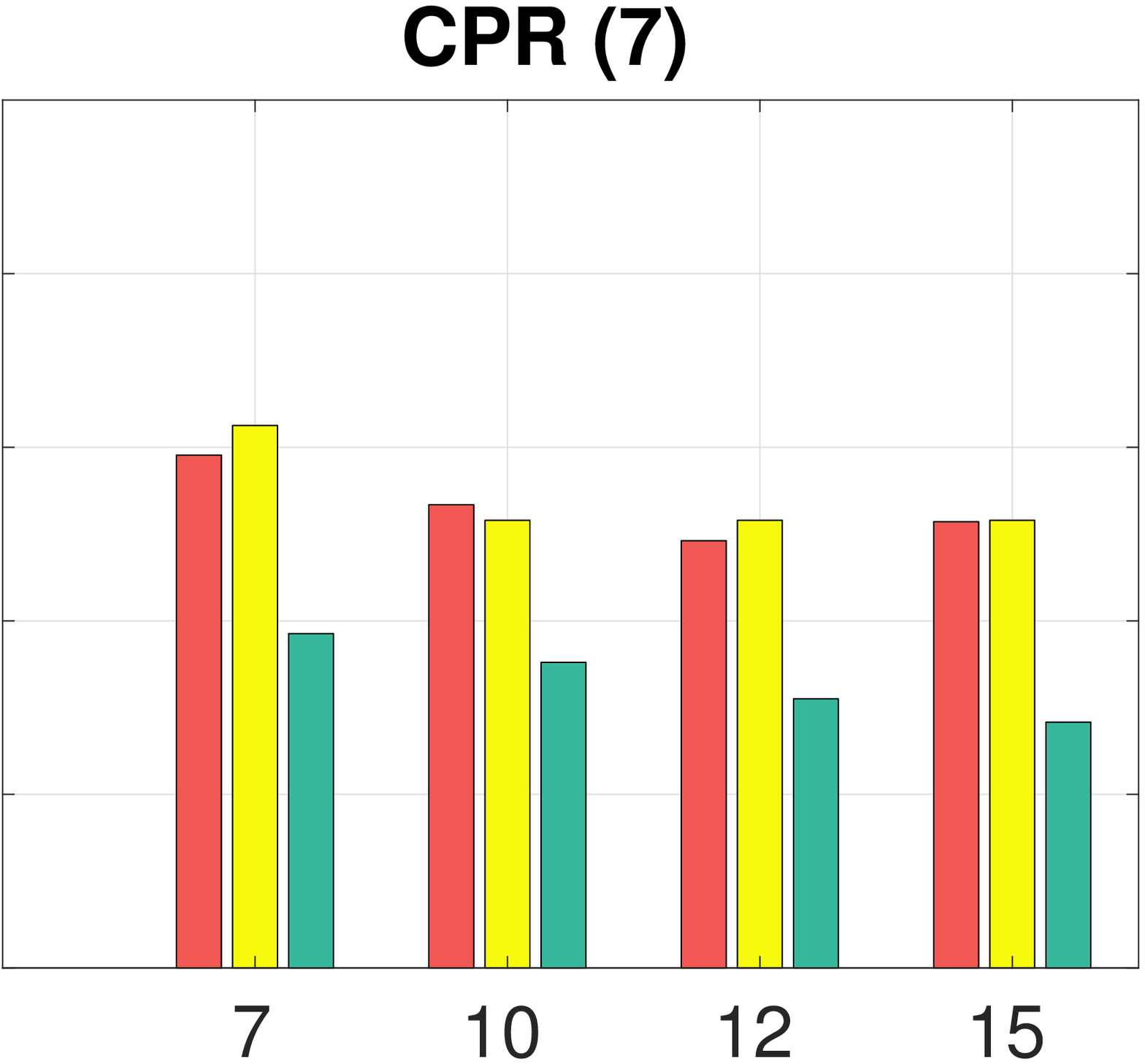}
	\includegraphics[height=2.8cm]{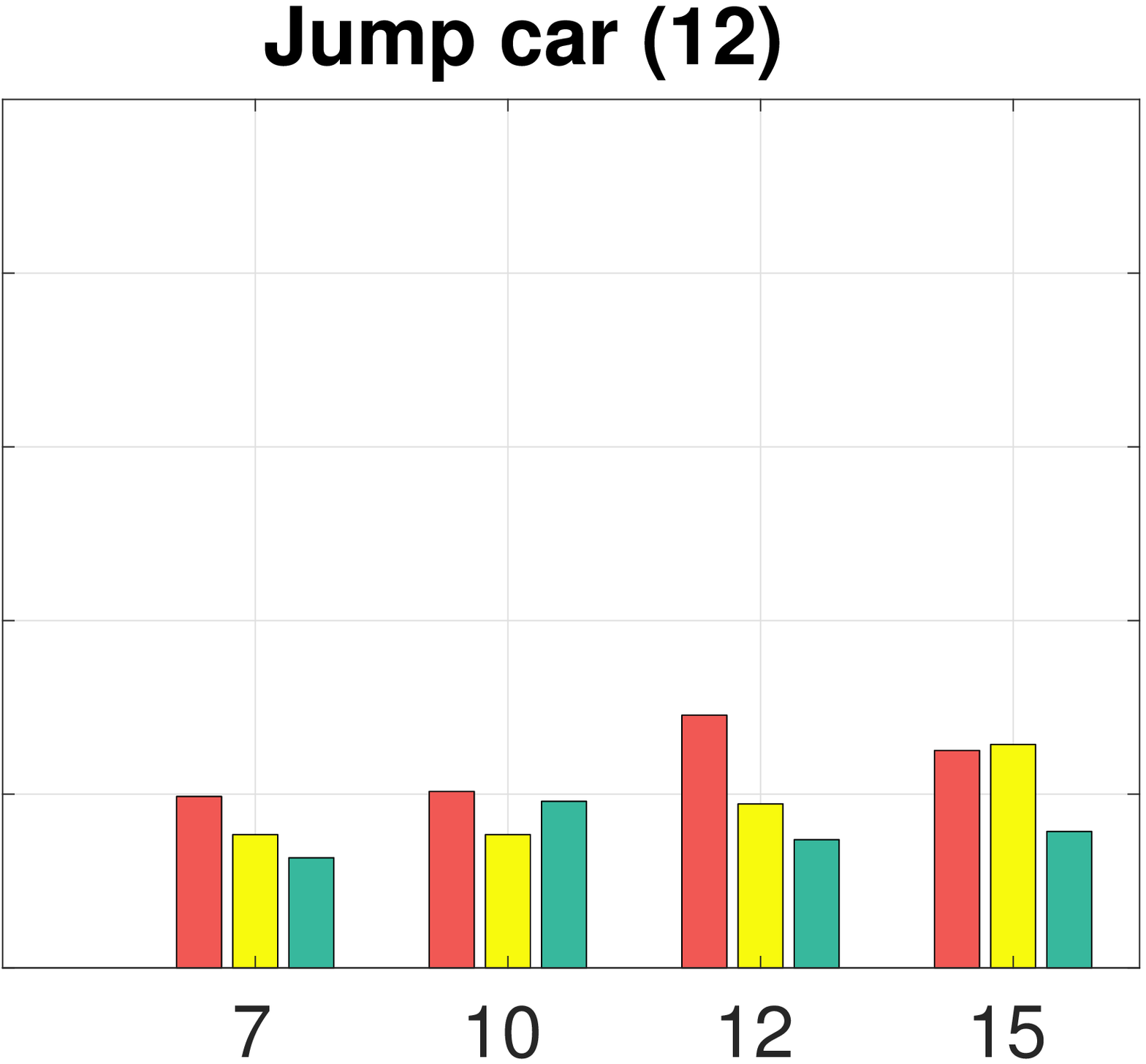}
	\includegraphics[height=2.8cm]{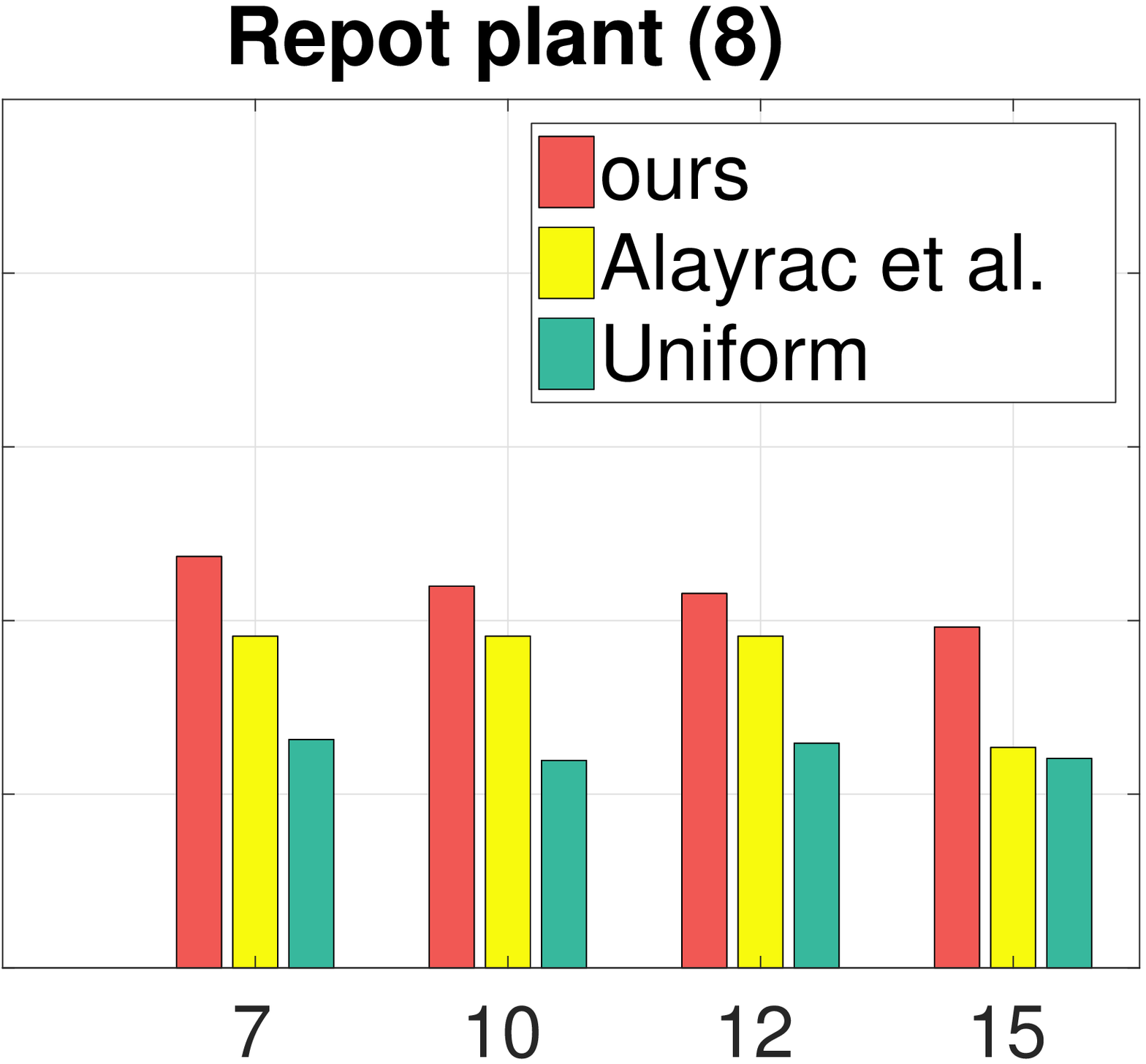}
\vspace{-0.2cm}
	\caption{Comparison of our method with Alayrac \etal~\cite{alayrac2016unsupervised} on the Instructional Videos Dataset~\cite{alayrac2016unsupervised}. To be compatible to the main step detection of Alayrac \etal~we report the mean over 15 randomly selected frames from each segment. }
\vspace{-0.2cm}
	\label{fig:inria_comparison} 
\end{figure*}

The only GMM parameter is $K$, the number of assumed sub-activities. We again consider Inria Instructional Videos without background and show the Mof as a function of $K$, once partially unsupervised (sub-activity appearance model from ground truth) and once fully unsupervised in Fig.~\ref{fig:inria_without_background}(a) and (b) respectively. As can be expected, the Mof drops when moving from the partially to the fully unsupervised case. This drop can be attributed to the fact that the Instructional Videos Dataset is extremely difficult, and exhibits a lot of variation across the videos. In both partially and fully unsupervised cases, however, the Mof remains stable with respect to $K$, demonstrating that our method is quite robust with respect to varying $K$. This is also the case once background is considered in the full model with the original sequences (see Fig.~\ref{fig:inria_comparison}). 

\subsection{Background Modelling}
In Fig.~\ref{fig:inria_standard_model_with_background}, we demonstrate the effectiveness of our full model in capturing the background in the original sequences in Inria Instructional Videos. Fig.~\ref{fig:inria_standard_model_with_background} shows the improvement in Mof once the background is accounted for in the model; there are improvements on every activity, with the most significant being a three-fold increase for \emph{`jump car'} despite the sequences being $83\%$ background. In Fig.~\ref{fig:segmentation_visual}, we show qualitative examples of how our model copes with background, where it succeeds, where it fails. 

\subsection{Comparison to State of the Art}
\begin{figure}[hb]
	\centering
	\begin{subfigure}[b]{0.48\columnwidth}
		\includegraphics[width=\columnwidth]{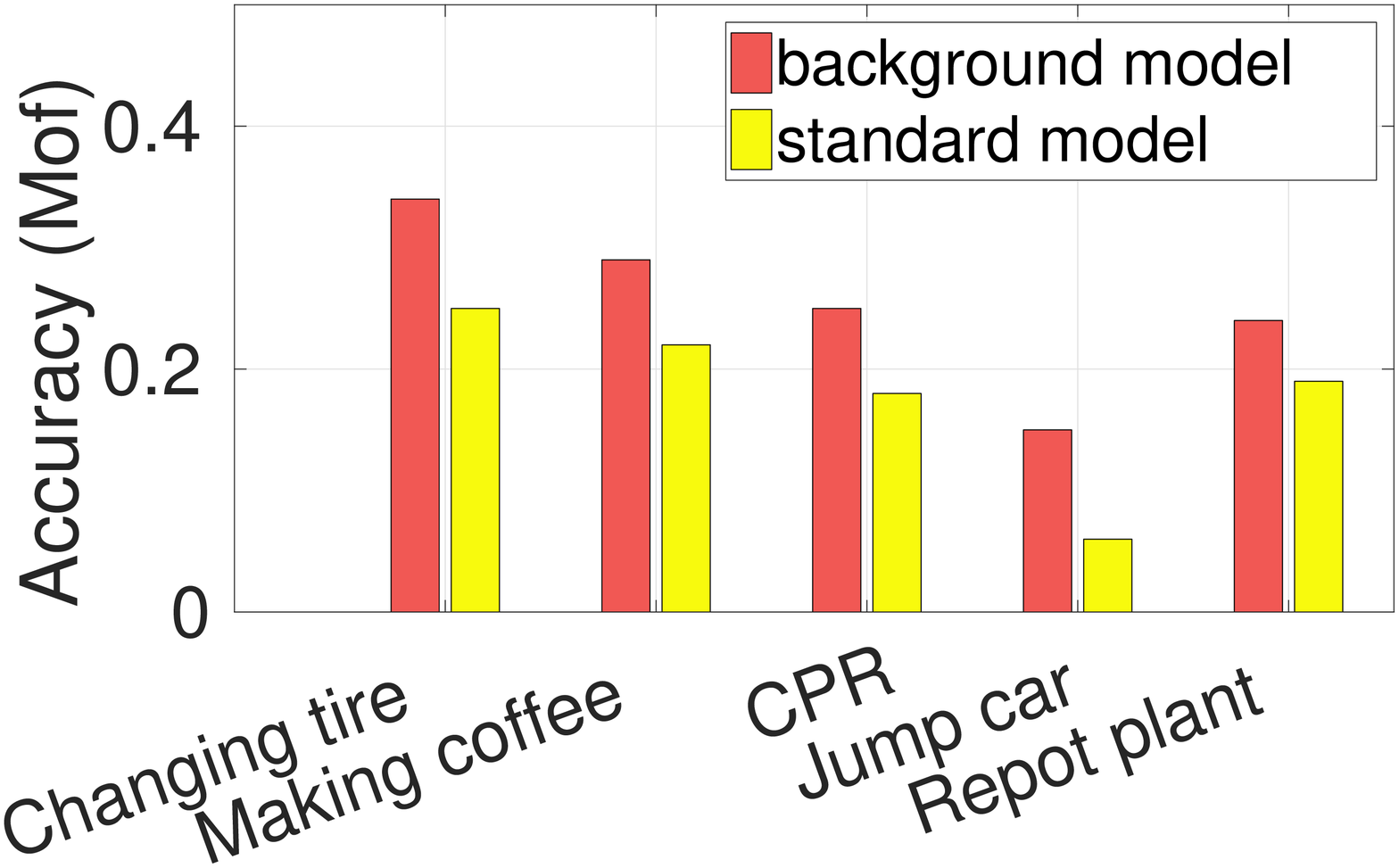}
		\caption{on ~\cite{alayrac2016unsupervised} }
		\label{fig:inria_standard_model_with_background}
	\end{subfigure}~ 
	\begin{subfigure}[b]{0.48\columnwidth}
	 \includegraphics[width=\columnwidth]{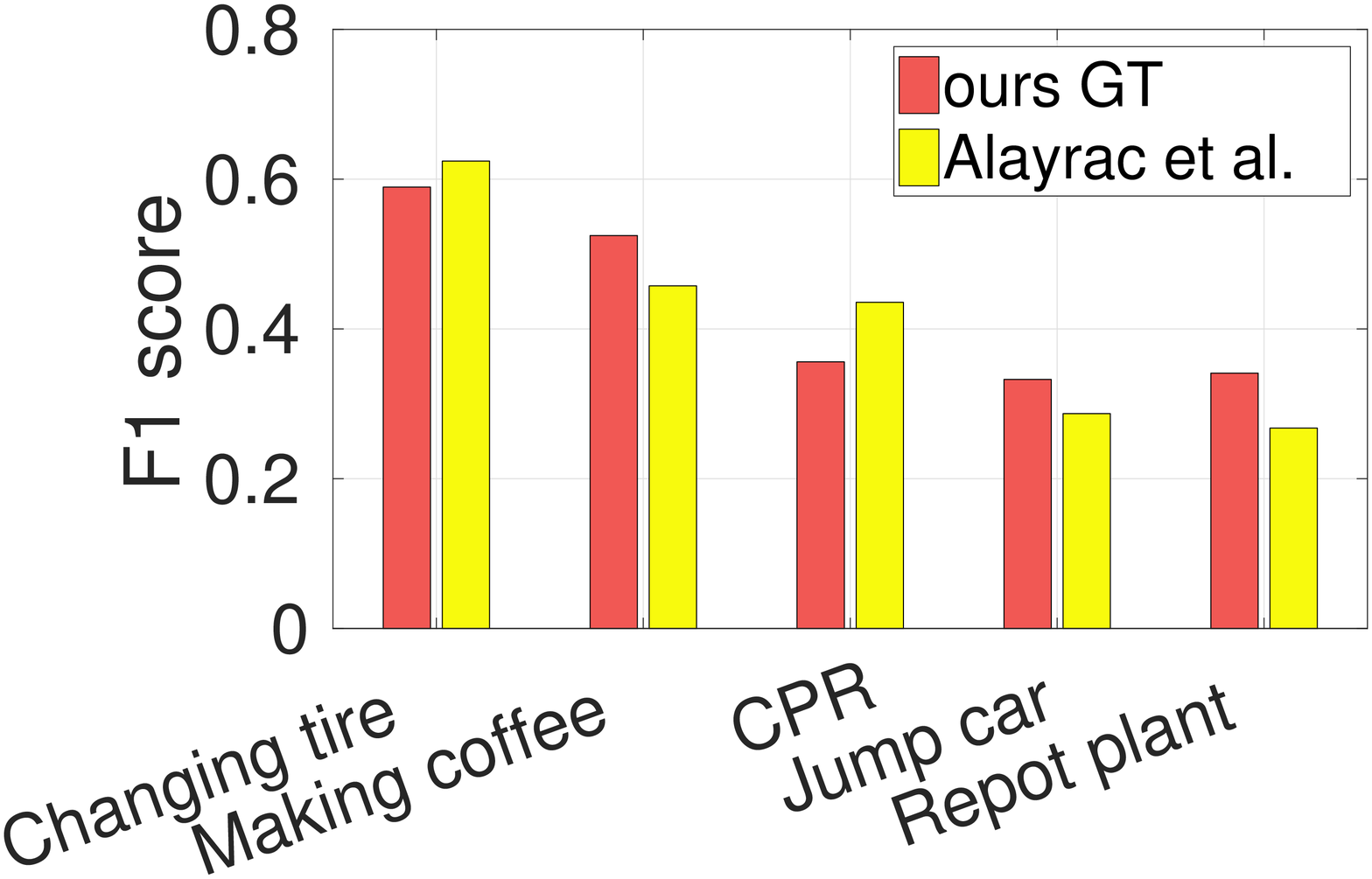}
  \caption{vs. Alayrac~\etal~\cite{alayrac2016unsupervised}}
  \label{fig:inria_comparison_GT}
 \end{subfigure}
\vspace{-0.2cm}
	\caption{ (a) Our standard model vs.\ background model on original Inria Instructional Videos sequences. The fractions of background are \emph{changing tire (0.46), making coffee (0.71), perform CPR (0.56), jump car (0.83) and repot plant (0.66)}. (b) Comparison of our supervised setting against Alayrac et al.'s supervised method on the Instructional Videos Dataset~\cite{alayrac2016unsupervised}. Here, our model learns the sub-activity appearance from the ground truth annotations. Alayrac \etal~use the ground truth annotations as constraints for their discriminative clustering based algorithm.}
	\label{fig:inria_standardwithBck_gtComp}
	\vspace{-0.3cm}
\end{figure}

\vspace{-0.1cm}
\paragraph{Inria Instructional Videos} 
We compare our full model to~\cite{alayrac2016unsupervised} in Fig.~\ref{fig:inria_comparison}, \ref{fig:inria_comparison_GT}. The method of~\cite{alayrac2016unsupervised} outputs a single representative frame for each sub-activity and reports an F1 score on this single frame. To make a valid comparison, since our work is aimed at finding entire segments, we randomly select a frame from each segment and then find a one-to-one mapping based on~\cite{liao2005clustering}. Our performance across the five activities is consistent and varies much less than~\cite{alayrac2016unsupervised}. We have stronger performance in three out of five activities, while we are worse on \emph{`perform cpr'} and \emph{`changing tire'}. The GMM is a distribution on permutations and orderings; it is by definition unable to account for repeating sub-activities but in \emph{`perform CPR'}, \emph{`give breath'} and \emph{`do compression'} are repeated multiple times and account for more than 50\% of the sequence frames.
In general, we attribute our stronger performance to the fact that the GMM can model flexible sub-activity orderings, while~\cite{alayrac2016unsupervised} enforces a strict ordering. The GMM parameter $\rhom$ has a prior with hyper-parameter $\rho_o$ (Sec.~\ref{sec:gmm}). A smaller $\rho_0$ allows more flexible orderings, while a larger $\rho_0$ encourages the ordering $\pim$ to remain similar to the canonical ordering $\sigmam$. In all of our reported results, we fixed $\rho_0\!\!=\!\!1$. We find that for an activity such as \emph{`change tire'}, which follows a strict ordering, a larger $\rho_0$ is more appropriate; with $\rho_0\!\!=\!\!5$ we are comparable to~\cite{alayrac2016unsupervised} (0.41 vs.\ 0.42 F1 score). 
For \emph{`jump car'} our method outperforms ~\cite{alayrac2016unsupervised}, however our overall performance is the lowest as our model struggles with separating the visually very similar \emph{`remove cable A'} and \emph{`remove cable B'}.

\begin{table} 
\centering
 \begin{tabular}{| r |r r r |} \hline
   &  & Mof & Jaccard \\ \hline
  \multirow{2}*{\small{Fully Supervised}} 
  & SVM~\cite{huang2016connectionist}  & 15.8 & -  \\ \cline{2-4}
  & HTK~\cite{kuehne2014language}   & 19.7 & -  \\ \cline{2-4} 
  \hline 
  
  \multirow{3}*{\small{Weakly Supervised}} 
  & OCDC~\cite{bojanowski2014weakly}   & 8.9 & 23.4 \\ \cline{2-4}
  & ECTC~\cite{huang2016connectionist}  & 27.7 & -  \\ \cline{2-4}
  & Fine2Coarse~\cite{richard2016temporal} & 33.3 & \textbf{47.3} \\ \hline
 
  \small{Unsupervised} & ours iterated      & \textbf{34.6} & 47.1 \\ \cline{2-4} \hline
 \end{tabular}
\vspace{-0.2cm}
 \caption{Comparisons on Breakfast Actions~\cite{kuehne2014language}. Methods are evaluated according to Mof and Jaccard index. For both, a higher result indicates better performance. 
 }
\label{table:breakfast_comparison}
\vspace{-0.3cm}
\end{table}

\vspace{-0.4cm}
\paragraph{Breakfast Actions} This dataset has no background labels so we apply our standard model and compare with other fully supervised and semi-supervised approaches in Table~\ref{table:breakfast_comparison}. Of the supervised methods, the SVM method~\cite{huang2016connectionist} classifies each frame individually without any temporal consideration and achieves an Mof of $15.8\%$. This shows the strength (and necessity) of temporal information. \emph{``Ours iterated''} is the only fully unsupervised method; we only set $K$ based on ground truth. In comparison, the weakly supervised methods~\cite{huang2016connectionist, richard2016temporal,bojanowski2014weakly} require both $K$ as well as an ordered list of sub-activities as input. ECTC~\cite{huang2016connectionist} is based on discriminative clustering, while OCDC~\cite{bojanowski2014weakly} and Fine2Coarse~\cite{richard2016temporal} are both RNN-based methods. We find that our fully unsupervised approach has performance that is state of the art. 

\section{Conclusion} 
In this paper we present an unsupervised method for partitioning complex activity videos into coherent segments of sub-activities. 
We learn a function assigning sub-activity scores to a video frame's visual features, we model the distribution over sub-activity permutations by a Generalized Mallows Model (GMM). Furthermore, we account for background frames not contributing to the actual activity. 

We successfully test our method on two datasets of this challenging problem and are either comparable to or out-perform the state of the art, even though our method is completely unsupervised, in contrast to the existing work. Our method is able to produce coherent segments, at the same time being flexible enough to allow missing steps and variations in ordering. Performance drops slightly for complex activities including repetitive sub-activities, as the GMM does not allow for such repeating structures. In the future we plan to investigate approaching this problem in a hierarchical manner to handle repeating blocks as a single step, which can then be further subdivided. Finally, the GMM is unimodal -- only one canonical ordering for the set is assumed. This is a valid assumption for activities such as cooking and simple procedural tasks, but we will consider for future work applying multi-modal extensions.

\vspace{-0.8cm}
\paragraph{Acknowledgments} Research in this paper was supported by the DFG project YA 447/2-1 (DFG Research Unit FOR 2535 Anticipating Human Behavior).

{\small
\bibliographystyle{ieee}
\bibliography{egbib}
}

\end{document}